\documentclass[preprints,Article,accept,moreauthors,pdftex]{Definitions/mdpi} 

\firstpage{1} 
\pubvolume{1}
\issuenum{1}
\articlenumber{0}
\pubyear{2022}
\copyrightyear{2022}
\externaleditor{\textls[-15]{Academic Editor: Andreas Holzinger}} 
\datereceived{3 December 2021} 
\dateaccepted{10 January 2022} 
\datepublished{13 January 2022} 
\hreflink{https://doi.org/} 
\usepackage{gensymb}
\usepackage{natbib}

\Title{A Transfer Learning Evaluation of Deep Neural Networks for Image Classification}

\TitleCitation{A Transfer Learning Evaluation of Deep Neural Networks for Image Classification}

\Author{{Nermeen Abou Baker} *\orcidA{}, Nico Zengeler \orcidB{} and Uwe Handmann \orcidC{}}
\AuthorNames{Nermeen Abou Baker, Nico Zengeler and Uwe Handmann}

\AuthorCitation{Abou Baker, N.; Zengeler, N.; Handmann, U.}

\address[1]{{Computer Science Institute, Ruhr West University of Applied Sciences, 46236 Bottrop}, Germany; Nico.Zengeler@hs-ruhrwest.de (N.Z.); uwe.handmann@hs-ruhrwest.de (U.H.)}

\corres{\hangafter=1 \hangindent=1.05em \hspace{-0.82em}Correspondence: nermeen.baker@hs-ruhrwest.de}

\abstract{Transfer learning is a machine learning technique that uses previously acquired knowledge from a source domain to enhance learning in a target domain by reusing learned weights. 
This technique is ubiquitous because of its great advantages in achieving high performance while saving training time, memory, and effort in network design. 
In this paper, we investigate how to select the best pre-trained model that meets the target domain requirements for image classification tasks. 
In our study, we refined the output layers and general network parameters to apply the knowledge of eleven image processing models, pre-trained on ImageNet, to five different target domain datasets. 
We measured the accuracy, accuracy density, training time, and model size to evaluate the pre-trained models both in training sessions in one episode and with ten episodes.}

\keyword{transfer learning; image classification; deep neural network} 

\begin{document}
\section{Introduction}
\label{Introduction}

Deep learning is a subfield of machine learning that allows computers to automatically interpret representations of data by learning from examples.
Transfer learning is a deep learning technique that uses previous knowledge to learn new tasks and is becoming increasingly popular in many applications with the support of Graphics Processing Unit (GPU) acceleration. 
Transfer learning has many benefits that have attracted researchers in different domains, to~name but a few: medical applications~\cite{Lundervold.2019}, remote sensing~\cite{PiresdeLima.2020}, optical satellite images~\cite{Zou.2018}, supporting automated recycling~\cite{S-Cube-BakSzaHan2020}, natural language processing~\cite{Houlsby.222019}, mobile applications~\cite{Choe.2020}, etc. 
However, there are some caveats in choosing the best pre-trained model for such applications, as~most focus on accuracy and leave out other important parameters. 
Therefore, it is important to also consider other metrics such as training time or memory requirements before proceeding to a concrete~implementation. 

Transfer learning is performed with pre-trained models, typically large Convolutional Neural Networks (CNNs) that are pre-trained on large standard benchmark datasets and then reused for the new target task. 
The reuse of such pre-trained models can be easily implemented by, for~example, replacing certain layers with other task-specific layers and then training the model for the target task. 
Moreover, many frameworks such as PyTorch, MATLAB, Caffe, TensorFlow, Onnx, etc., provide several pre-trained models that can help researchers implement this promising technique. 
The state-of-the-art has many architectures, each with its own characteristics, that are suitable for CNN applications. 
However, the~performance of the resulting transfer learning network depends on the pre-trained model used. 
Before going into the reuse of these models, it seems that there is a great deal of freedom in choosing the~model.

According to~\cite{DBLP:journals/corr/abs-1811-01533}, the~size and similarity of the target dataset and the source task can be used as rules of thumb to choose the pre-trained model. ImageNet is a leading dataset due to its popularity and data diversity. However, fine-tuning pre-trained models that are trained on ImageNet is not per se able to achieve good results on spectrograms, for~example.
Besides, following the previous strategy might not be enough with the current challenging constraints that require high accuracy, a short training time, and~limited hardware resources for specific applications.
Previously pre-trained model analysis was presented in~\cite{Canziani.2016}, who collected reported values from the literature and compared the models' performance on ImageNet to evaluate several scores, such as the top-five accuracy normalized to model complexity and power~consumption.

Another worthwhile attempt was presented by~\cite{Bianco.2018}, who benchmarked pre-trained models on ImageNet using multiple indices such as accuracy, computational complexity, memory usage, and~inference time to help practitioners better fit the resource~constraints.

Choosing the best pre-trained model is a complex dilemma that needs to be well understood, and~researchers could feel confused about picking the most suitable option. 
We performed extensive experiments to classify five datasets on eleven pre-trained models. We provide in-depth insight and offer a feasible guideline for transfer learning that uses a pre-trained model by introducing an overview of the tested models and datasets and evaluating their performance using different metrics.
Since most pre-trained models are used to classify ImageNet, we conducted our research on different datasets, including standard and non-standard~tasks.

The paper is organized as follows: It starts by introducing the research gap in the Introduction in Section~\ref{Introduction}. Section~\ref{Summary of related learning methods} summarizes the related learning methods.
Section~\ref{Materials and Methods} gives an overview of the main characteristics of the tested models and datasets. 
Section~\ref{Implementation} focuses on the implementation of the models. 
Results are presented and discussed in Section~\ref{Results}. 
Finally, the~conclusion of the work is given in Section~\ref{Conclusions}.

\section{Summary of Related Learning~Methods}\label{Summary of related learning methods}
 Machine learning is data-hungry; therefore, it has tremendous success in data-intensive applications, but~it is limited when the dataset is small. This section summarizes different types of related machine learning methods for solving image classification tasks, including zero-shot learning, one-shot learning, few-shot learning, and~transfer learning. One common advantage of these methods is that they leave out the burden of collecting large-scale supervised data and the issue of data~scarcity.
 
 \subsection{Zero-Shot~Learning}
 With zero-shot learning, it is possible to train a model without accessing data with non-observed labels during training by using previous labels and some auxiliary information. It assumes that the model can classify instances of unseen visual examples. This method looks promising when new unlabeled examples are introduced frequently~\cite{socher2013zeroshot}.
 In the zero-shot learning method, the~test set and training class set are disjoint~\cite{xian2020zeroshot}. Several solutions to this problem have been proposed, such as learning intermediate attribute classifiers~\cite{6571196}, learning a mixture of seen class proportions~\cite{zhang2015zeroshot}, or~compatibility learning frameworks~\cite{2016}, for~example.
 
 \subsection{One-Shot~Leaning}
 One of the limitations of deep learning is that it demands a huge amount of training data examples to learn the weights. However, one-shot learning seeks to predict the required output based on one or a few learning examples. However, this is usually achieved by either sharing feature representations~\cite{1467333} or model parameters~\cite{NIPS2004_ef1e491a}. Methods such as this are useful for classification tasks when it is hard to classify data for every possible class or when new classes are added~\cite{socher2013zeroshot}. One-shot learning has been proven to be an efficient method as the number of known labels grows because in this case, it is most likely that the model has already learned a label that is very similar to the one to be learned~\cite{Tommasi2009TheMY}.
 
 \subsection{Few-Shot~Learning}
 This method refers to feeding a model with a small number of training data samples. It is useful for applications that lack information or can be accessed only with difficulty due to concerns about privacy, safety, or~ethical issues~\cite{10.1145/3386252}.
 
 \subsection{Transfer~Learning}
 In line with the previously mentioned methods, and~according to~\cite{azadi2017multicontent,liu2019feature,luo2017label}, transfer learning methods often use few-shot learning, where prior knowledge is transformed from the source task into a few-shot task~\cite{azadi2017multicontent}. There are two ways to implement transfer learning: fine-tuning only the classifier layers, which keeps the entire model's weight constant, excluding the last layer, and~fine-tuning all layers, which allows the weights to change throughout the entire network.~Section~\ref{Implementation} describes these two ways technically~\cite{8560550}.
\section{Models and~Datasets}\label{Materials and Methods}

In this section, we present the CNN-design-based architectures as a critical factor in constructing the pre-trained models, the~tested models, and~the~datasets.

\subsection{CNN-Design-Based~Architectures}\label{CNN design}
The CNN is the fundamental component in developing a pre-trained model, and to understand the architecture, some criteria define the design architecture of the models, as~follows:

\begin{itemize}
 \item Depth: The NN depth is represented by the number of successive layers. Theoretically, deep NNs are more efficient than shallow architectures, and~increasing the depth of the network by adding hidden layers has a significant effect on supervised learning, particularly for classification tasks~\cite{Montufar.282014}. However, cascading layers in a Deep Neural Network (DNN) is not straightforward, and~this may cause an exponential increase in the computational cost;
 \item Width: The width of a CNN is as significant as the depth. Stacking layers may learn various feature representations, but~they would not learn useful features. Therefore, a DNN should be wide enough, so the loss at the local minima could be smaller with larger layer widths~\cite{Kawaguchi.2019};
 \item Spatial kernel size: A CNN has many parameters and hyperparameters, including weights, biases, the number of layers, the activation function, the learning rate, and~the kernel size, which define the level of granularity. Choosing the kernel size affects the correlation of neighboring pixels. Smaller filters extract local and fine-grained features, whereas larger filters extract coarse-grained features~\cite{Khan.2020};
 \item Skip connection: Although a deeper NN yields better performance, it may face challenges in performance degradation, vanishing gradients, or~higher test and training errors~\cite{Hochreiter.1998}. To~tackle these problems, the shortcut layer connection was first proposed by~\cite{Srivastava.532015} by skipping some intermediate layers to allow the special flow of information across the layers, for~example zero-padding, projection, dropout, skip connections, etc;
 \item Channels: CNNs have powerful performance in learning features automatically, and~this can be dynamically performed by tuning the kernel weights. However, some feature maps have little or no role in object discrimination~\cite{Hu.2018} and could cause overfitting as well. Those feature maps (or the channels) can be optimally selected in designing the CNN to avoid overfitting.
\end{itemize}

\subsection{Neural Network~Architectures}
\begin{figure}[H]
\includegraphics[width=0.9\textwidth]{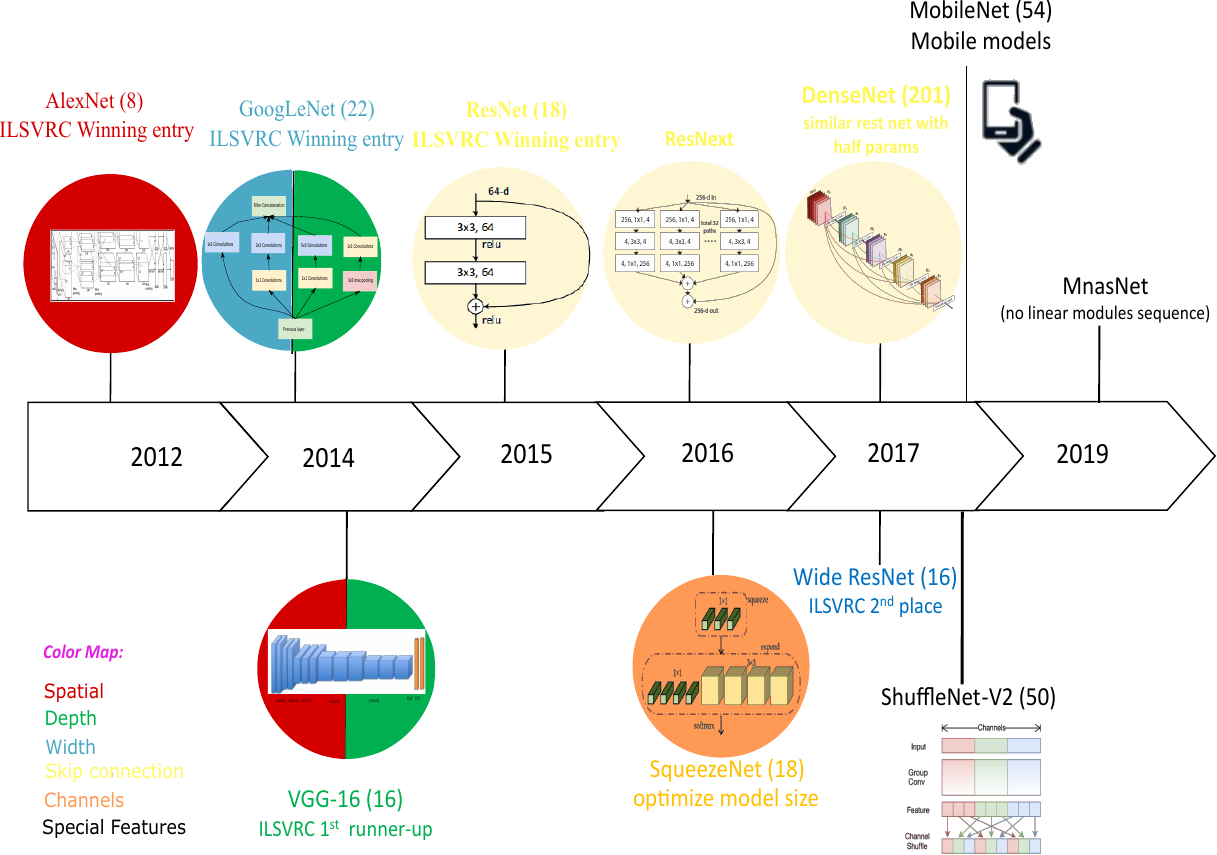}
\caption{Infographic of the tested pre-trained models. Each model is introduced with its architecture symbol, the~number of layers between brackets, and~design specification (see the color map).\label{fig1_info}}
\end{figure}

This study tested eleven popular pre-trained models. Figure~\ref{fig1_info} gives a comprehensive infographic representation over time. Table~\ref{tab1} depicts all the tested models with their main characteristics based on their design, which is discussed in Section~\ref{CNN design}.

\begin{table}[H]
\small
\caption{ The tested models with their main characteristics, where * refers to features specially designed for the~model. \label{tab1}}
\newcolumntype{C}{>{\centering\arraybackslash}X}
\begin{tabularx}{\textwidth}{ccccc}
\toprule
\textbf{Model}	& \textbf{Year}	& \textbf{Depth} & \textbf{Main Design Characteristics} & \textbf{Reference}\\
\midrule
AlexNet	&2012	&8	&Spatial	&\cite{AlexKrizhevsky.2012}
\\
VGG-16	&2014	&16	& Spatial and depth	&\cite{Simonyan.942014}
\\
GoogLeNet	&2014	&22	&Depth and width	&\cite{Szegedy.6720156122015}\\
ResNet-18	&2015	&18	&Skip connection	&\cite{He.12102015}\\
SqueezeNet	&2016	&18	&Channels	&\cite{Iandola.2242016}\\
ResNext	&2016	&101	&Skip connection	&\cite{Xie.72120177262017}\\
DenseNet	&2017	&201	&Skip connection	&\cite{Huang.72120177262017}\\
MobileNet	&2017	&54	&Depthwise separable conv *	&\cite{Howard.4172017}\\
WideResNet	&2017	&16	&Width	&\cite{Zagoruyko.5232016}\\
ShuffleNet-V2	&2017	&50	&Channel shuffle * &\cite{Zhang.742017}\\
MnasNet	&2019	&No linear sequence	&Neural architecture search *	&\cite{Tan}\\
\bottomrule
\end{tabularx}
\end{table}
\unskip

\unskip

\subsection{Datasets}
A combination of standard datasets was tested, which were: CIFAR10 with 60 K images~\cite{Zaheer}, Modified National Institute of Standards and Technology (MNIST) with 70 K images~\cite{Kaziha.1119201911212019}, Hymenoptera~\cite{NEURIPS2019_9015}, and~non-standard, which were: smartphones and augmented smartphones~\cite{smartphones.EAI}, as~follows:
 
 \subsubsection{Hymenoptera}
This is a small RGB dataset that is used to classify ants and bees from a PyTorch tutorial on transfer learning. 
It consists of $245$ training images and $153$ testing~images.

\subsubsection{Smartphone~Dataset}
\begin{figure}[H]
\includegraphics[width=0.6\textwidth]{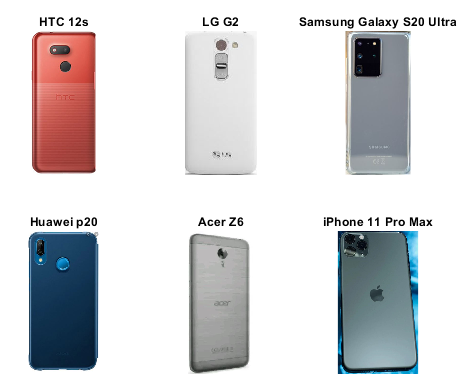}
\caption{Example of a subset of the smartphone~dataset.\label{fig5_smartphone}}
\end{figure}

This is a relatively small dataset of different smartphone models, representing six brands, namely: Acer, HTC, Huawei, Apple, LG, and~Samsung. 
It contains 654 RGB images with twelve classes, which are: Acer Z6, HTC 12S, HTC R70, Huawei Mate 10, Huawei P20, iPhone 5, iPhone 7 Plus, iPhone 11 Pro Max, LG G2, LG Nexus 5, Samsung Galaxy S20 Ultra, and~Samsung S10E. We created this dataset as a case study in a previous work~\cite{smartphones.EAI}, to~show that transfer learning can reach high accuracy with a small dataset to support automated e-waste recycling through device classification.
We collected the images from the search engines focusing on the backside where unique features such as the logo and camera lenses, which are distinguishing because most front-sides of modern smartphones look similar, as~showcased in Figure~\ref{fig5_smartphone}.

\unskip

\subsubsection{Augmented Smartphone~Dataset}
Data augmentation is usually used to increase the volume of the dataset effortlessly. We applied a rotation operation in combination with increasing the noise. 
For the rotation operation, we rotated by $r \in \{$45\degree$,~$135\degree$,~$225\degree$,~$315\degree$\}$; for~the noise operation, we added noise in percentages $p \in \{10\%,~25\%,~50\%\}$ by adding pixels from a discrete uniform distribution $\{0 … 255 \cdot p\}$. 
This resulted in a total of twelve augmentation operations for each image. Therefore, the~total number of images was multiplied by $12$ to obtain a total number of $8502$ images in the augmented dataset, including the $654$ original~images.

\section{Implementation}\label{Implementation}
This study performed two scenarios under the same condition, using an Nvidia GTX 1080 TiGPU to train and evaluate eleven PyTorch vision models in a sequential fashion, namely AlexNet, VGG-16, Inception-V1 (GoogLeNet), ResNet-18, SqueezeNet, DenseNet,
ResNext, MobileNet, Wide ResNet, ShuffleNet-V2, and~MnasNet. We re-trained each model on five tasks, namely MNIST, CIFAR10, Hymenoptera, smartphones, and~augmented smartphones, each in a grid search over learning rates $\eta \in \{10^{-2}, 10^{-3}, 10^{-4}\}$ with the ADAM optimizer and a batch size equal to 10. 
In our plots, we show only the model with the highest accuracy in the overall learning rate. 
To overcome overfitting, we performed early stopping, so we saved model weights only if the validation accuracy increased. That is, if~the validation accuracy decreased, we still used the best model found so~far.

We chose to perform two experiments in our paper where a pre-trained model was used~to: 
\begin{itemize}
 \item Fine-tune the classifier layer only: This method keeps the feature extraction layers from the pre-trained model fixed,~so-called frozen. We then re-initialized the task-specific classifier parts, as~given by reference in the PyTorch vision model implementations~\cite{NEURIPS2019_9015}, with~random values. If~the PyTorch model did not have an explicit classifier part, for~example the~ResNet18 architecture, we fine-tuned only the last fully connected layer. We froze all other weights during training. This technique saved training time and, to some degree, overcame the problem of a small-sized target dataset because it only updated a few weights;
 \item Fine-tune all layers: For this method, we used the PyTorch vision models with original weights as pre-trained on ImageNet and fine-tuned the entire parameter vector. In~theory, this technique achieves higher accuracy and generalization, but~it requires a longer training time since it is used for initializing weights by continuing the backpropagation instead of random initialization in scratch training.
\end{itemize}

PyTorch vision models typically have a classifier part and a feature extraction part. Fine-tuning the output layers means fine-tuning the classifier part, which results in a large variation in the model size. We froze all other weights during training.
We assessed the model performance with four metrics: the accuracy, the~accuracy density, the~model size, and~training time on a~GPU.
\subsection{Accuracy~Density}
This represents the accuracy divided by the number of parameters:
\begin{equation}
 density = \frac{accuracy}{\#parameters}
\end{equation}

\textls[-15]{A higher value corresponds to a higher model efficiency in terms of parameter~usage. }

\subsection{Accuracy and Model Sizes vs. Training~Time}
Along with measuring accuracy across tasks, we also measured the training time in seconds and the number of learning parameters in MB. The~more complex the model is, the~more parameters need to be optimized. 
When determining the memory utilization of a GPU for each model, the~number of parameters is critical. This is the amount of memory that will be allocated to the network and the amount of memory needed to process a~batch.

\section{Results}\label{Results}

We present our results for two experiments, learning from one episode and learning from ten episodes. 
In each experiment, we tested the fine-tuning of both the classifier batch and the entire network. 
In the configurations with few shots, each sample was presented only once in a single training episode, while in the configuration with ten episodes, each sample was presented ten times~accordingly.

\subsection{One-Episode~Learning}

\subsubsection{Fine-Tuning the Full~Layers}
As shown in Figure~\ref{Acc-DenOne}, we calculated the average accuracy densities of all tested datasets, and~we found that SqueezeNet with full tuning showed the highest accuracy density among all models, particularly AlexNet, which came in tenth place. This result affirmed the original hypothesis when SqueezeNet was designed, that it preserves AlexNet's accuracy with $50$-times fewer parameters and less than a $0.5$ MB model size~\cite{Iandola.2242016}.

\subsubsection{Fine-Tuning the Classifier Layers~Only}
\begin{figure}[H]
\includegraphics[width=0.6\textwidth]{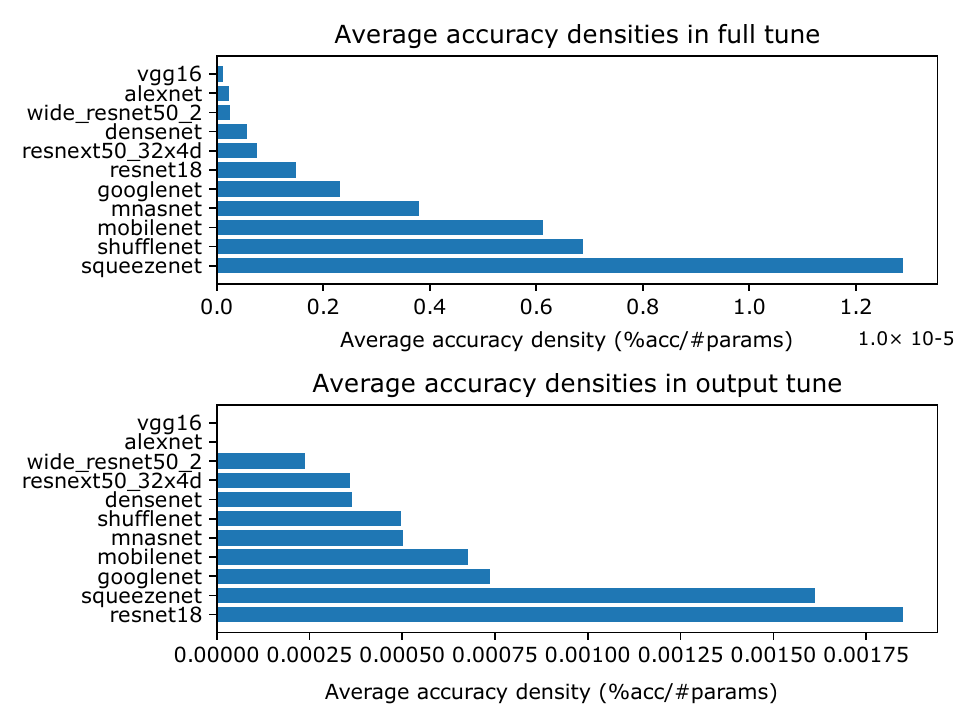}
\caption{{Average accuracy densities with full tuning and tuning the classifier layer only for one~episode.} 
\label{Acc-DenOne}} 
\end{figure}

The results, as~seen in Figure~\ref{Acc-DenOne}, were slightly different in terms of the accuracy density in the order of the models, but it showed a big difference in the values, where ResNet18 was the most suitable~candidate. Each Dataset is tested for both experiments and shown in detail in Appendix \ref{AppAccDenOne}.

\unskip

\subsection{Ten-Episode~Learning}
We tested 10 independent trials and calculated the average results to avoid any bias, as~follows.

\subsubsection{Fine-Tuning the Full~Layers}
Figure~\ref{Acc-DenTen} shows that ten-episodes learning did not affect the order of the models in terms of the accuracy densities compared with the one-episode experiments, and~the values were higher to a small degree.
Figure~\ref{FullBbl} shows the average model sizes and accuracy vs. the training time for all tasks and models after fine-tuning the full layers of all~datasets.

\subsubsection{Fine-Tuning the Classifier Layers~Only}
\begin{figure}[H]
\includegraphics[width=0.6\textwidth]{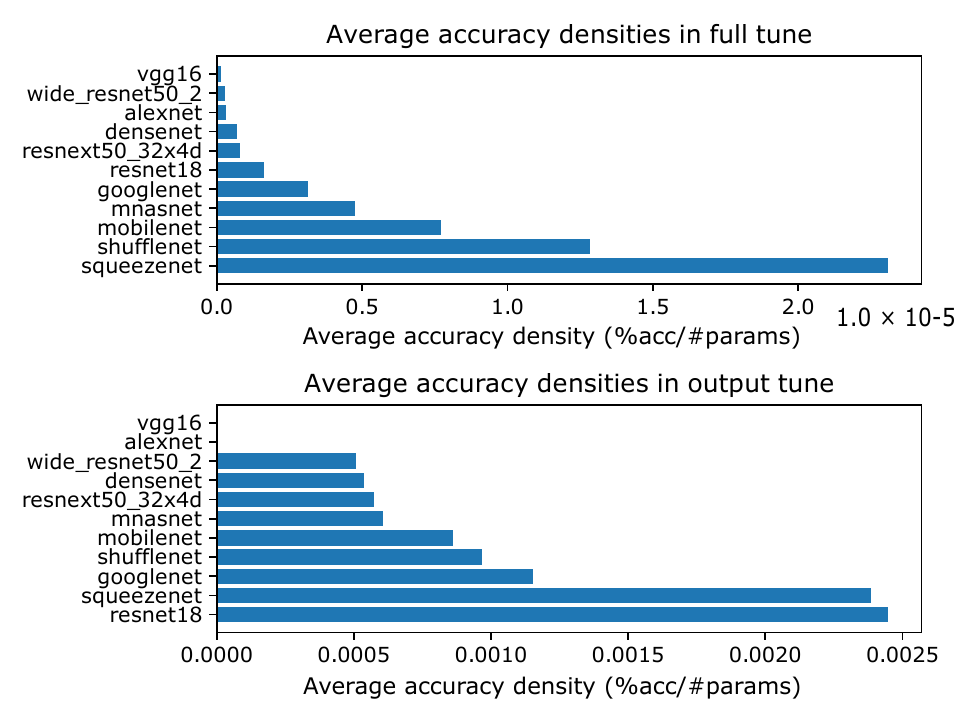}
\caption{{Average accuracy densities with full tuning and tuning the classifier layer only for ten~episodes}.\label{Acc-DenTen}} 
\end{figure}

We found that ResNet18 showed a satisfactory result again as the most efficient model that used its parameters efficiently, as~shown in Figure~\ref{Acc-DenTen}. Figure~\ref{classonly} shows the average model sizes and accuracy vs. the training time for all tasks and models after fine-tuning the classifier layer only of all datasets.
MnasNet had the poorest performance, making it the least-favorable model in terms of the error metrics, yet it showed a low model complexity and a short training time. The~most complex model in all experiments was VGG-16, and~the accuracy density figures confirmed this fact. As~a result, it might be less trustworthy for embedded and mobile~devices. Each Dataset is tested for both experiments and shown in detail in Appendix \ref{AppAccDenTen}.

\begin{figure}[H]
\includegraphics[width=0.7\textwidth]{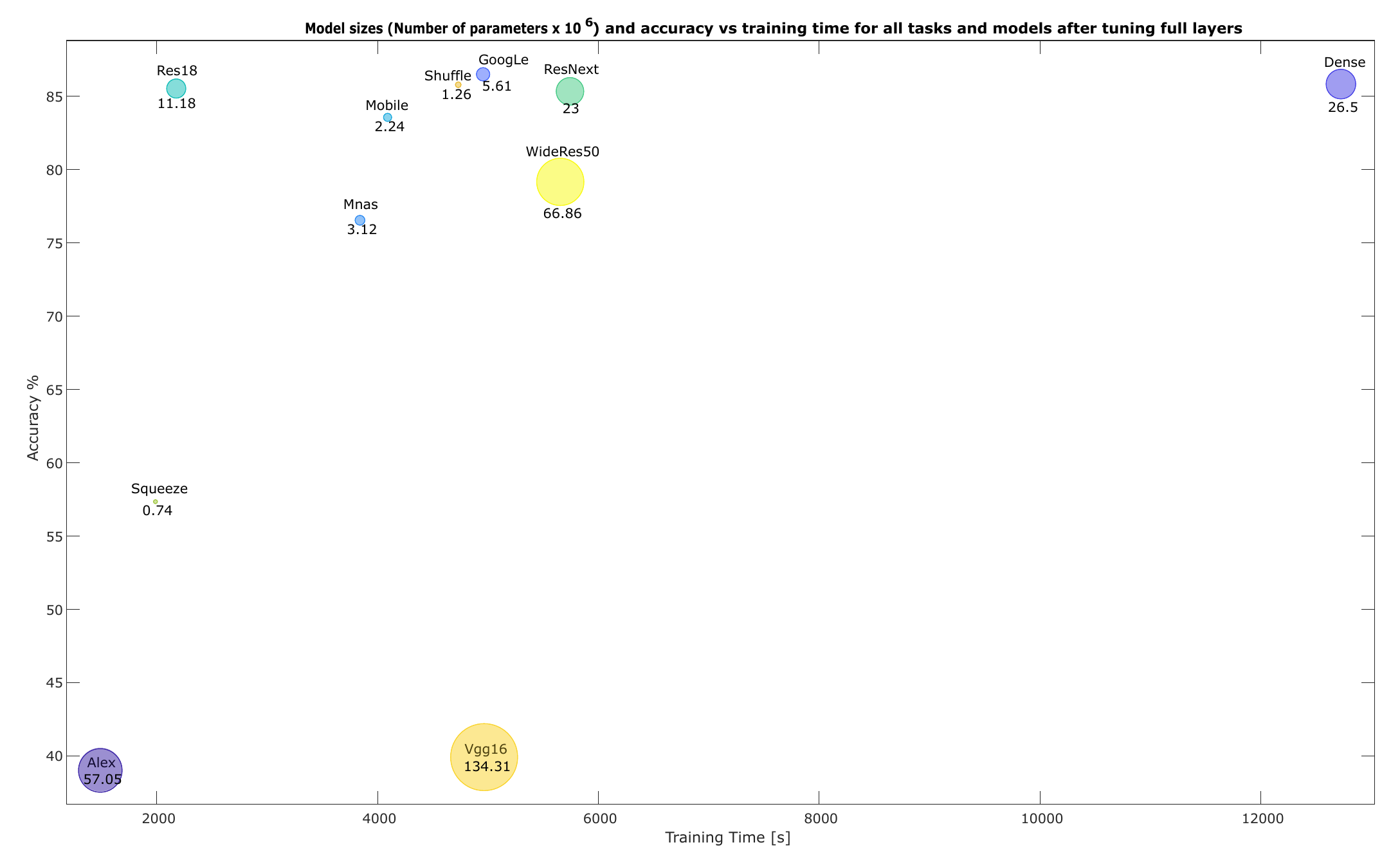}
\caption{Model sizes and accuracy vs. training time for all tasks and models after fine-tuning full~layers. \label{FullBbl}}
\end{figure}

\begin{figure}[H]
\includegraphics[width=0.7\textwidth]{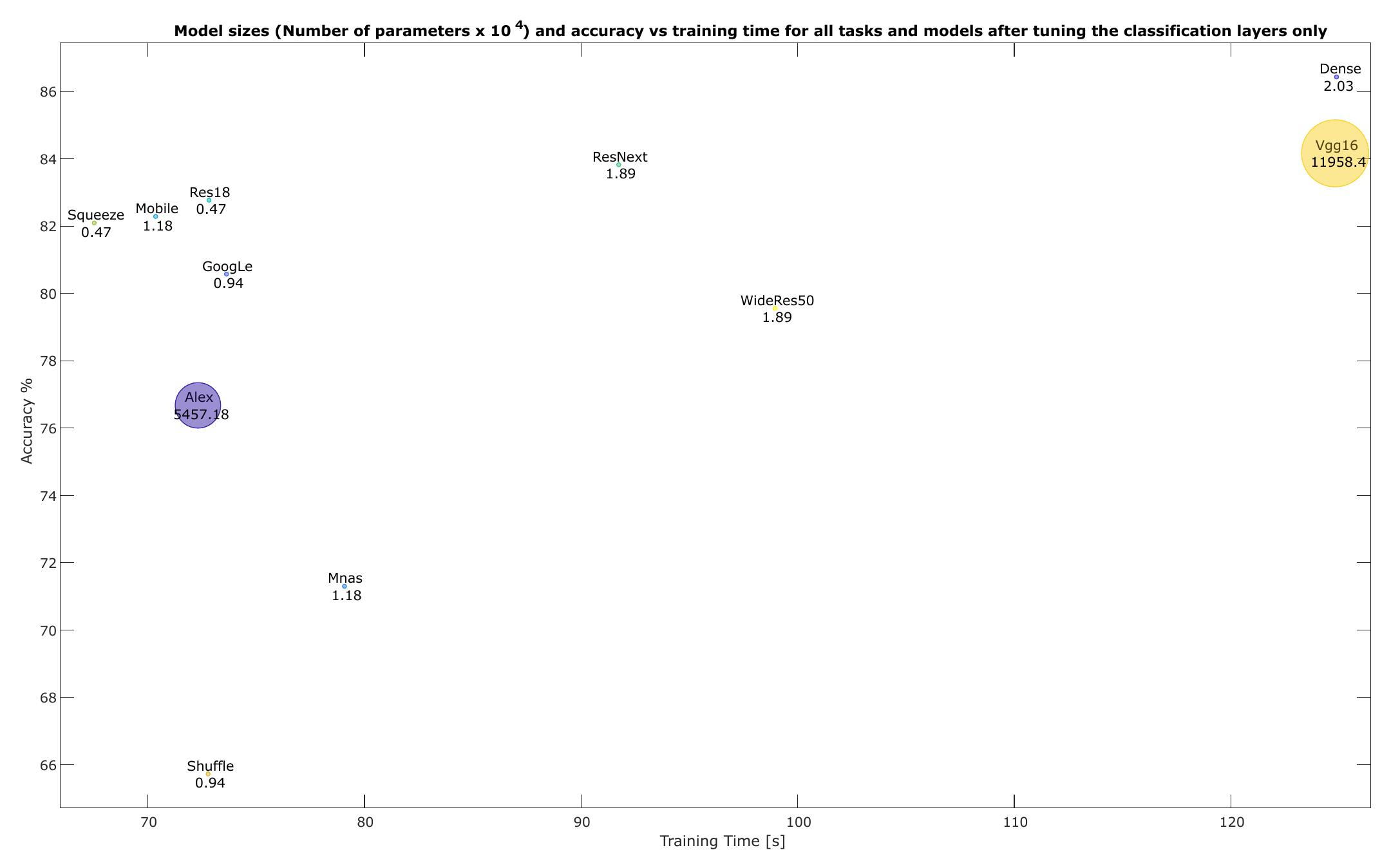}
\caption{Model sizes and accuracy vs. training time for all tasks and models after fine-tuning the classifier layer~only.\label{classonly}}
\end{figure}

Tuning hyperparameters means finding the best set of parameter values for a learning algorithm.
In CNNs, the~initial layers are designed to extract global features, whereas the later ones are more task-specific. Therefore, when tuning the classification layer, only the final layer for classification is replaced, while the other layers are frozen (the weights of the other layers are fixed). This means utilizing the knowledge of the overall architecture as a feature extractor and using it as a starting point for retraining. Consequently, it achieved high performance with a smaller number of parameters and a shorter training time, as~shown in Figure~\ref{classonly}. Usually, this scenario is used when the target task labels are scarce~\cite{rs13224712}. On~the other hand, full tuning means retraining the whole network (the weights are updated after each epoch) with a longer training time and more parameters, as~shown in Figure~\ref{FullBbl}. When target task labels are plentiful, this scenario is typically~applied.
Each Dataset is tested for tuning the classifier layer only and tuning full layers and shown in detail for ten episodes in Appendix \ref{AppAccModTen}, and for one episode in Appendix \ref{AppAccModOne}.

\unskip

\nointerlineskip

\section{Conclusions}\label{Conclusions}

DNNs' performance has been enhanced over time in many aspects. Nonetheless, there are critical parameters that define which pre-trained model perfectly matches the application requirements. 
In this paper, we presented a comprehensive evaluation of eleven popular pre-trained models on five datasets as a guiding tool for choosing the appropriate model before deployment. We conducted two different sets of experiments: one-episode learning and ten-episode learning, with~each experiment involving tuning the classifier layer only and full tuning. 
The previous findings, however, might provide some clues for choosing the right model for fine-tuning the classification layer only. For~applications that require high accuracy, GoogLeNet, DenseNet, ShuffleNet-V2, ResNet-18, and ResNext are the best candidates, while SqueezeNet is for the accuracy density, and AlexNet for the shortest training time, and~SqueezeNet, ShuffleNet, MobileNet, MnasNet, and~GoogLeNet are almost equal regarding the smallest model size, for embedded systems applications, for~example. On~the other hand, we can also provide some suggestions when fine-tuning only the classification layers. DenseNet achieved the highest accuracy, while ResNet18 the best accuracy density, and SqueezeNet the shortest training time. In~addition, all models had small model sizes except AlexNet and VGG-16.
Although we provided guidelines and some hints, our argumentation does not give a final verdict, but~it supports decisions for choosing the right pre-trained model based on the task~requirements.

Thus, for~specific application constraints, selecting the right pre-trained model can be challenging due to the tradeoffs among training time, model size, and~accuracy as decision factors to produce better~scores.

For future work, we plan to test more evaluation metrics with the provided parameters to facilitate decision-making in choosing the optimum model to fine-tune. Furthermore, we aim to systematically investigate the usability of all available a priori and a posteriori metadata for estimating useful transfer learning~hyperparameters.

\vspace{6pt} 

\authorcontributions{Conceptualization, N.A.B. and N.Z.; methodology, N.A.B.; software, N.Z.; writing---original draft preparation, N.A.B.; writing---review and editing, N.A.B. and N.Z.; supervision, U.H. All authors have read and agreed to the published version of the~manuscript.}

\funding{This work has been partially funded by the Ministry of Economy, Innovation, Digitization, and~Energy of the State of North Rhine-Westphalia within the project~Prosperkolleg.}

\institutionalreview{ Not applicable.}

\informedconsent{Not applicable.}

\dataavailability{No new data were created or analyzed in this study. Data sharing is not applicable to this article.}

\conflictsofinterest{The authors declare no conflict of~interest.} 

\appendixtitles{yes} 
\appendixstart
\appendix
\section[\appendixname~\thesection]{\highlighting{Accuracy Densities for Each Task with One-Episode~Learning}\label{AppAccDenOne} 
}
\begin{figure}[H]
\includegraphics[width=\textwidth]{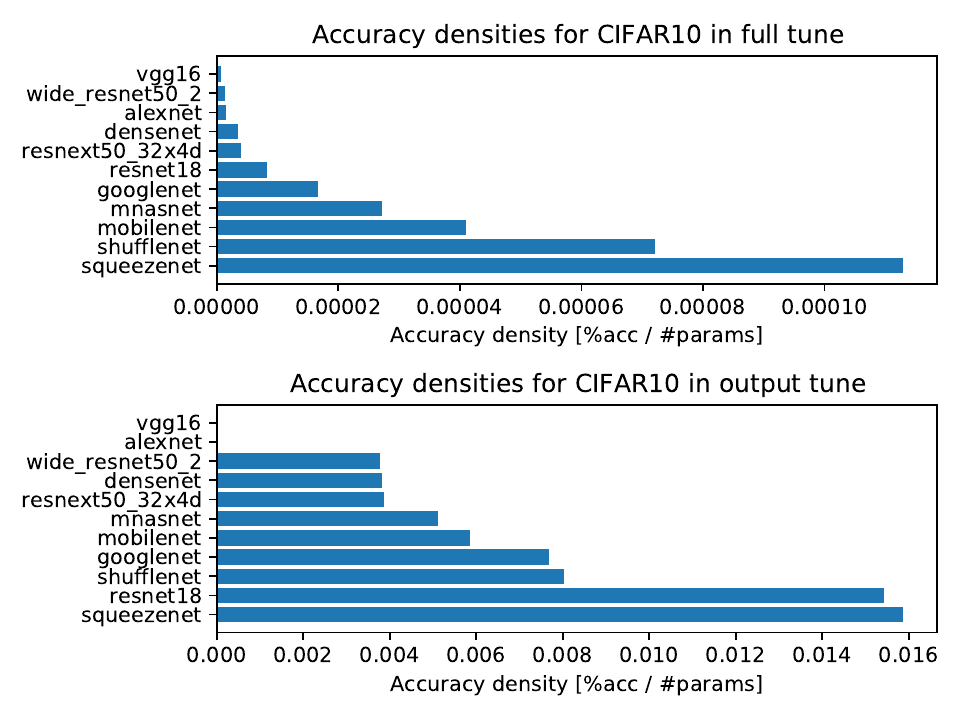}
\caption{Accuracy densities for one-episode learning on~CIFAR-10.}
\label{oneshot_cifar}

\end{figure}

\begin{figure}[H]

\includegraphics[width=\textwidth]{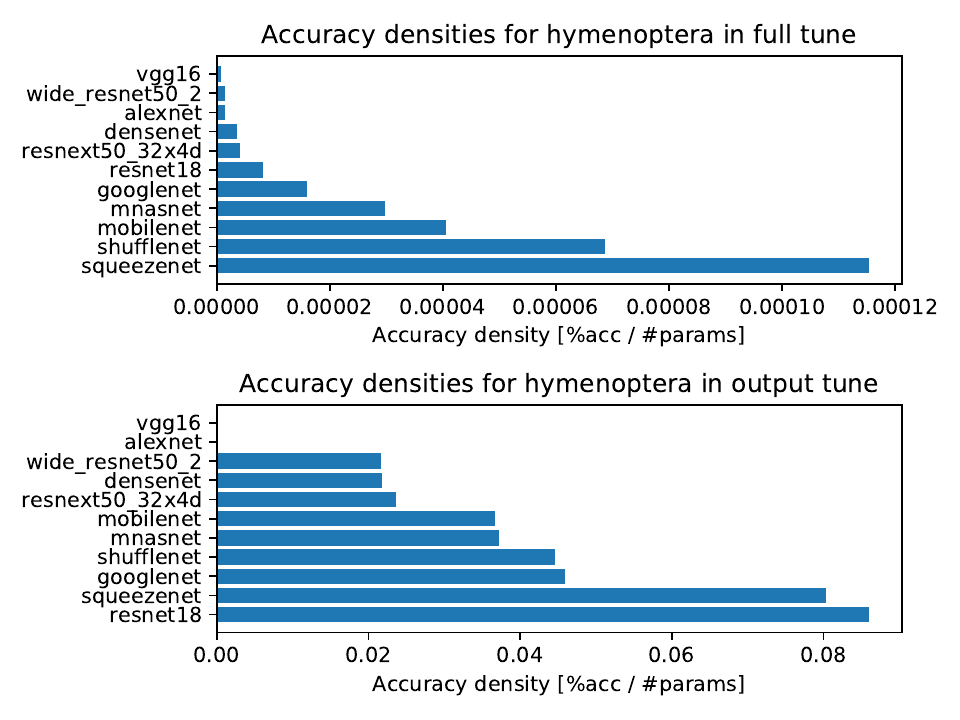}
\caption{Accuracy densities for one-episode learning on~Hymenoptera.}
\label{oneshot_hymenoptera}

\end{figure}

\begin{figure}[H]
\includegraphics[width=\textwidth]{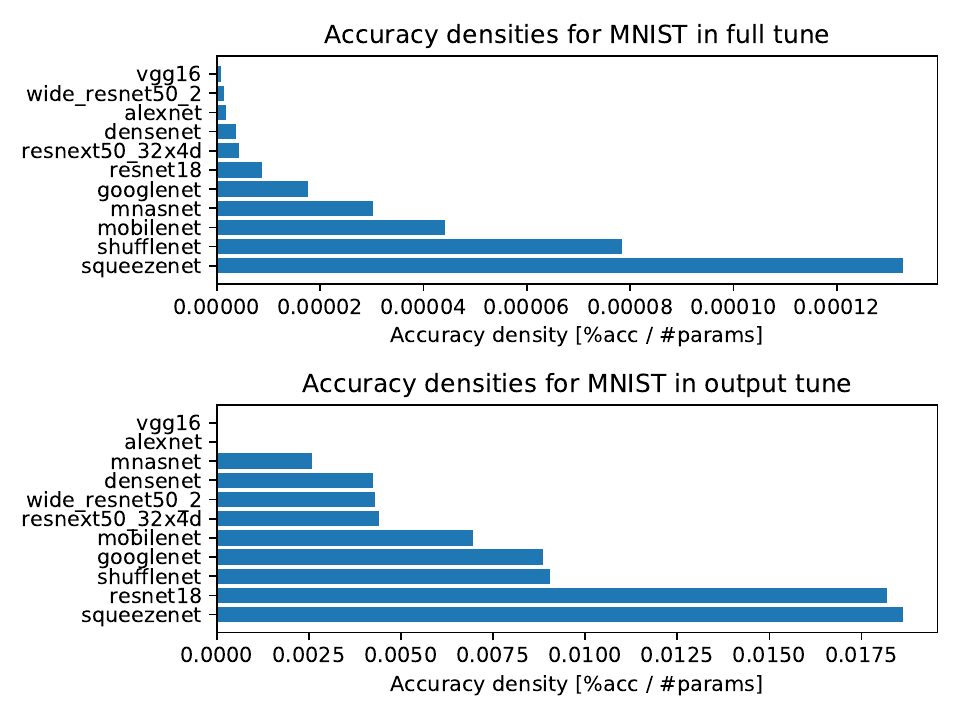}
\caption{Accuracy densities for one-episode learning on~MNIST.}
\label{oneshot_mnist}
\end{figure}

\begin{figure}[H]
\includegraphics[width=\textwidth]{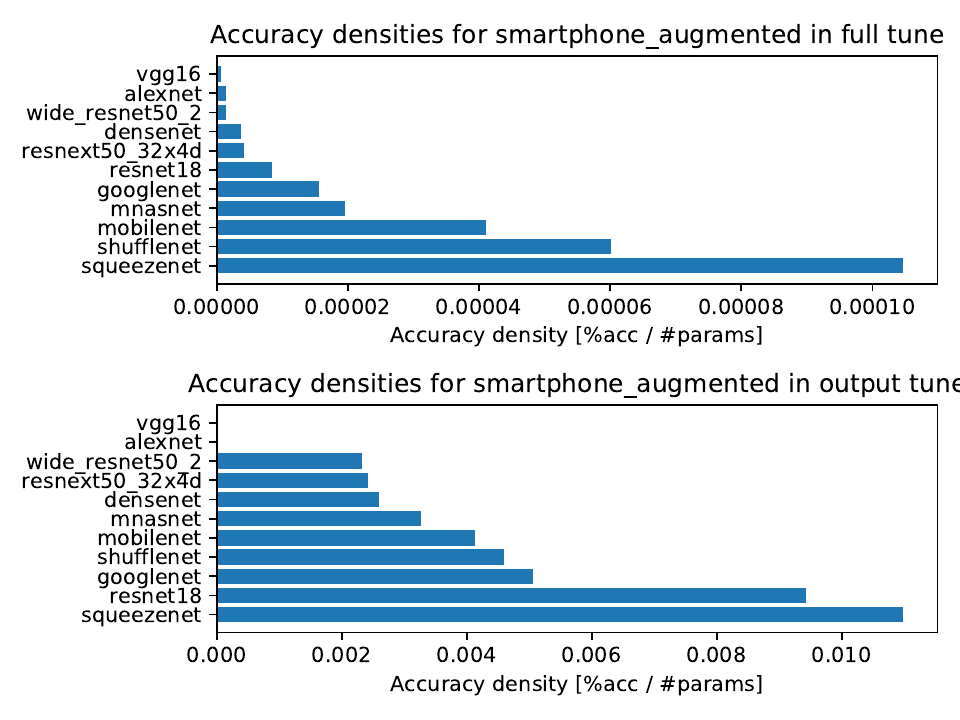}
\caption{Accuracy densities for one-episode learning on augmented smartphone~data.}
\label{oneshot_smartphone_augmented}
\end{figure}

\begin{figure}[H]
\includegraphics[width=\textwidth]{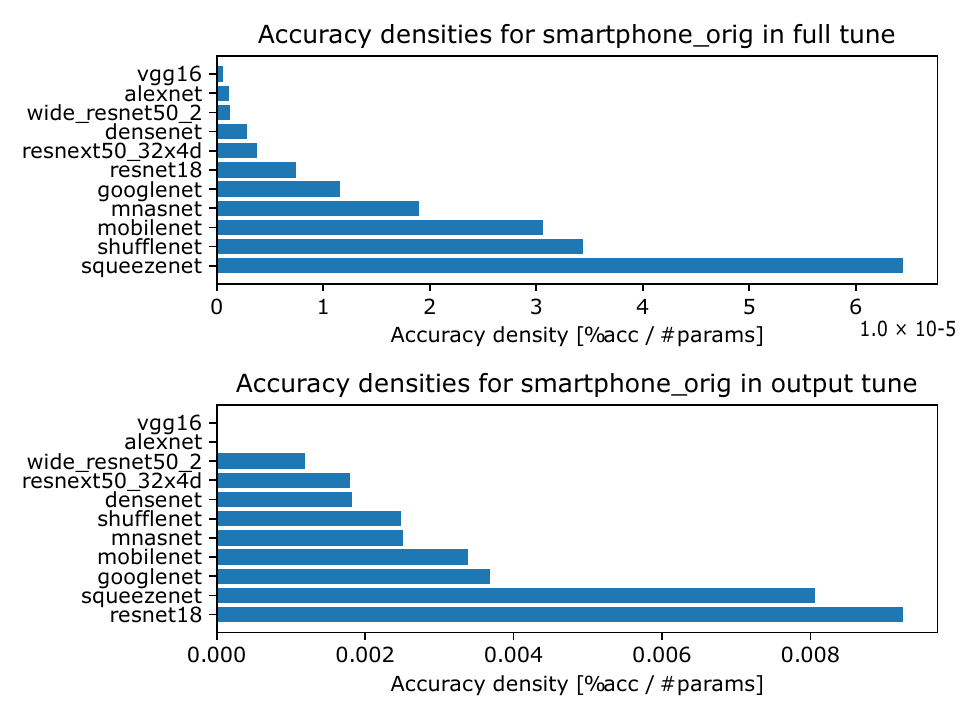}
\caption{{Accuracy} densities for one-episode learning on original smartphone~data.}
\label{oneshot_smartphone_orig}
\end{figure}

\begin{itemize}
\item Accuracy densities for one-episode learning on CIFAR-10, Figure \ref{oneshot_cifar};
 \item Accuracy densities for one-episode learning on Hymenoptera, Figure \ref{oneshot_hymenoptera};
\item Accuracy densities for one-episode learning on MNIST, Figure 
 \ref{oneshot_mnist};
\item \textls[-15]{Accuracy densities for one-episode learning on augmented smartphone data, Figure~\ref{oneshot_smartphone_augmented};}
\item Accuracy densities for one-episode learning on original smartphone data, Figure~\ref{oneshot_smartphone_orig}.
\end{itemize}
\unskip

\unskip

\unskip

\unskip

\section[\appendixname~\thesection]{Accuracy Densities for Each Task with Ten-Episode~Learning }\label{AppAccDenTen}
\begin{figure}[H]
\includegraphics[width=\textwidth]{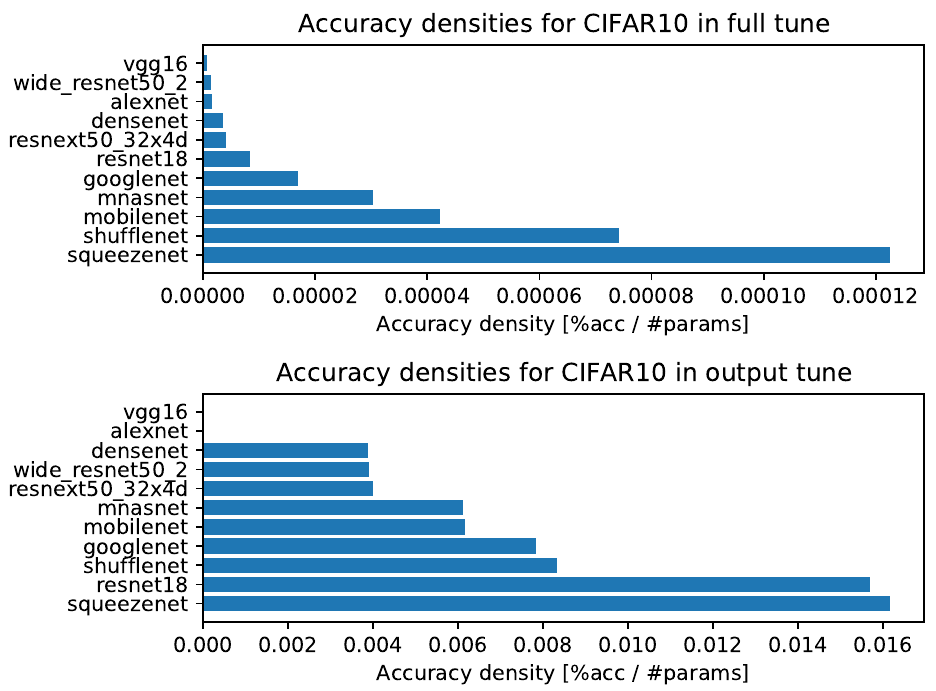}
\caption{Accuracy densities for ten-episodes learning on~CIFAR-10.}
\label{10ep_cifar}
\end{figure}

\begin{figure}[H]
\includegraphics[width=\textwidth]{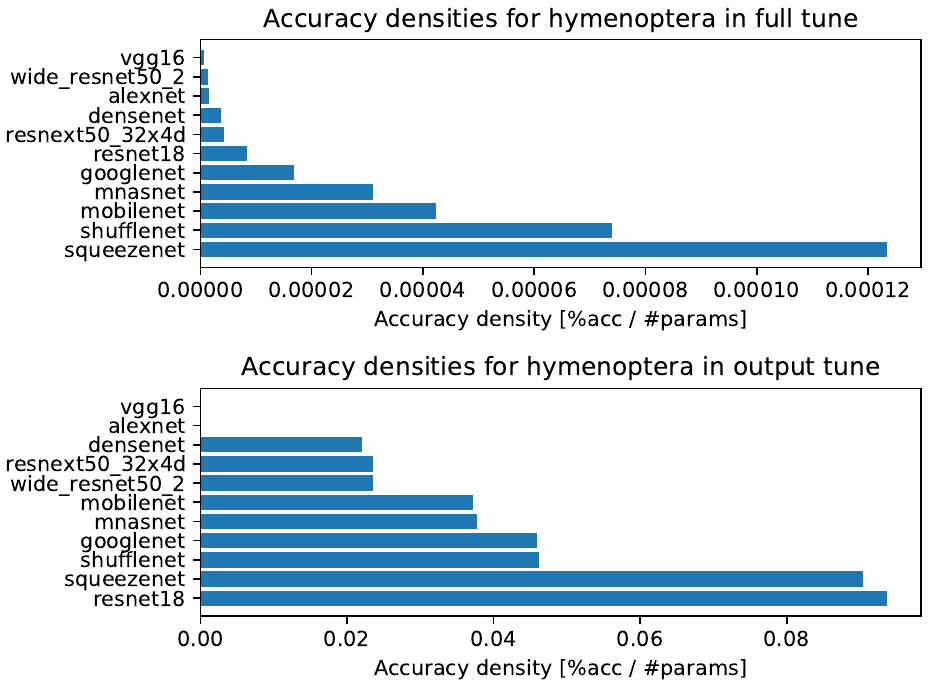}
\caption{Accuracy densities for ten-episodes learning on~Hymenoptera.}
\label{10ep_hymenoptera}
\end{figure}

\begin{figure}[H]
\includegraphics[width=\textwidth]{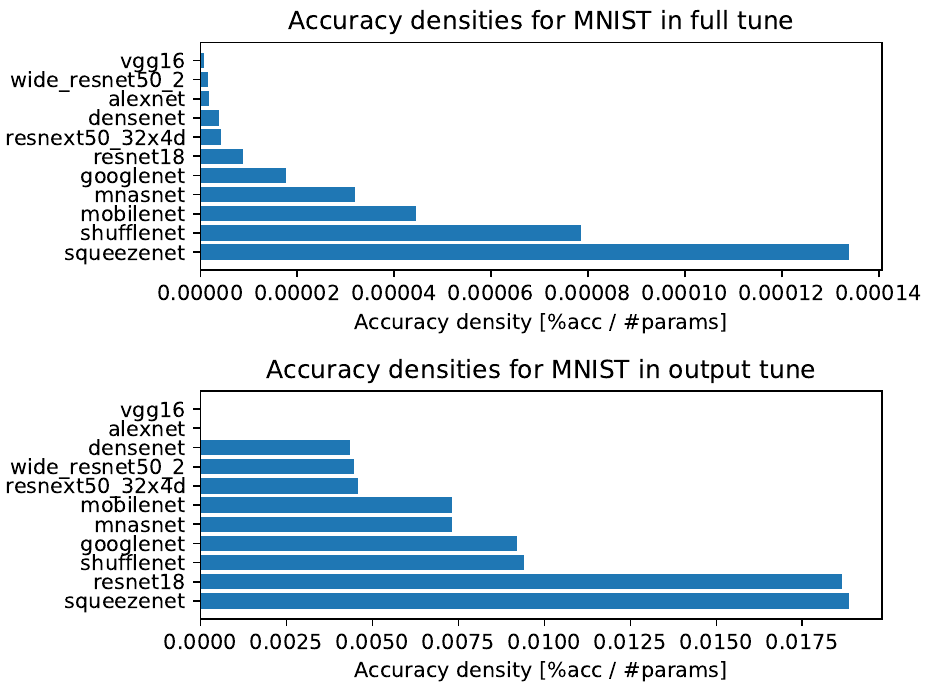} 
\caption{Accuracy densities for ten-episodes learning on~MNIST.}
\label{10ept_mnist}
\end{figure}

\begin{figure}[H]
\includegraphics[width=0.98\textwidth]{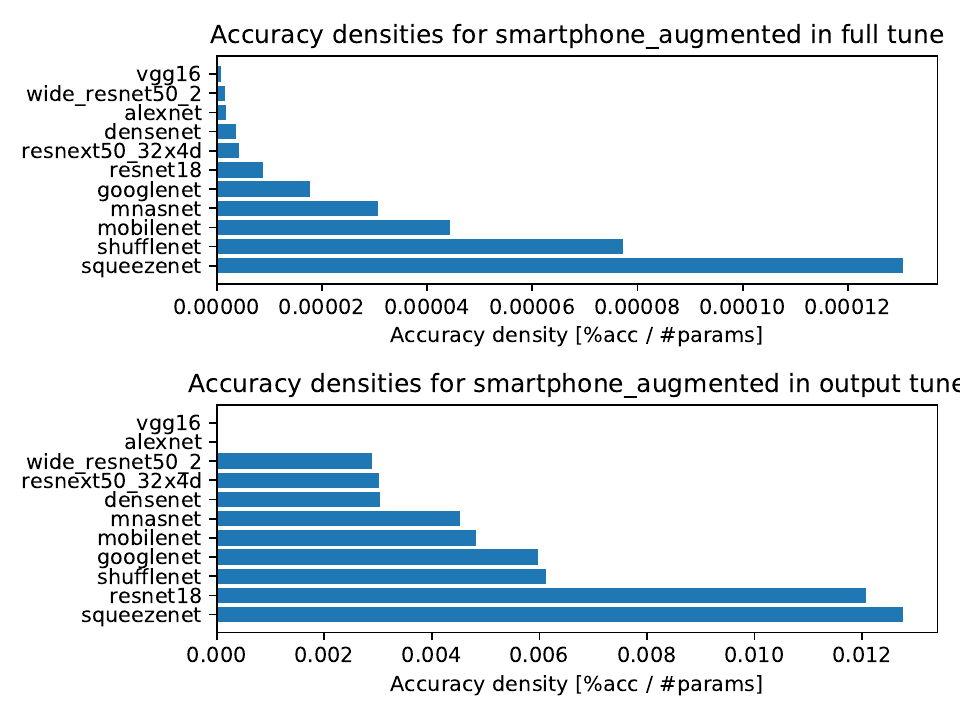}
\caption{Accuracy densities for ten-episodes learning on augmented smartphone~data.}
\label{10ep_smartphone_augmented}
\end{figure}

\begin{figure}[H]
\includegraphics[width=0.98\textwidth]{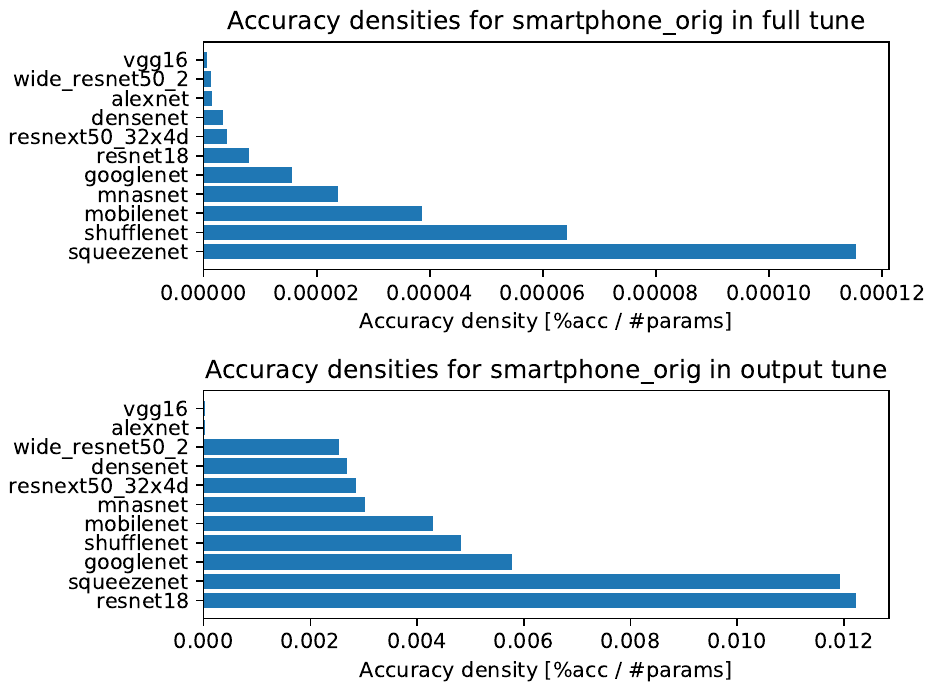}
\caption{Accuracy densities for ten-episodes learning on original smartphone~data.}
\label{10ep_smartphone_orig}
\end{figure}

\begin{itemize}
\item Accuracy densities for ten-episode learning on CIFAR-10, {Figure} \ref{10ep_cifar}; 
 \item Accuracy densities for ten-episode learning on Hymenoptera, Figure \ref{10ep_hymenoptera};
\item Accuracy densities for ten-episode learning on MNIST, Figure 
 \ref{10ept_mnist};
\item \textls[-5]{Accuracy densities for ten-episode learning on augmented smartphone data, \mbox{Figure~\ref{10ep_smartphone_augmented};}}
\item Accuracy densities for ten-episode learning on original smartphone data, Figure \ref{10ep_smartphone_orig}.
\end{itemize}
 
\unskip

\unskip

\unskip

\unskip

\section[\appendixname~\thesection]{Accuracy vs. Training Time and Number of Parameters (Model Size) for Each Task with Ten-Episode Learning}\label{AppAccModTen}
\begin{figure}[H]
\includegraphics[width=\textwidth]{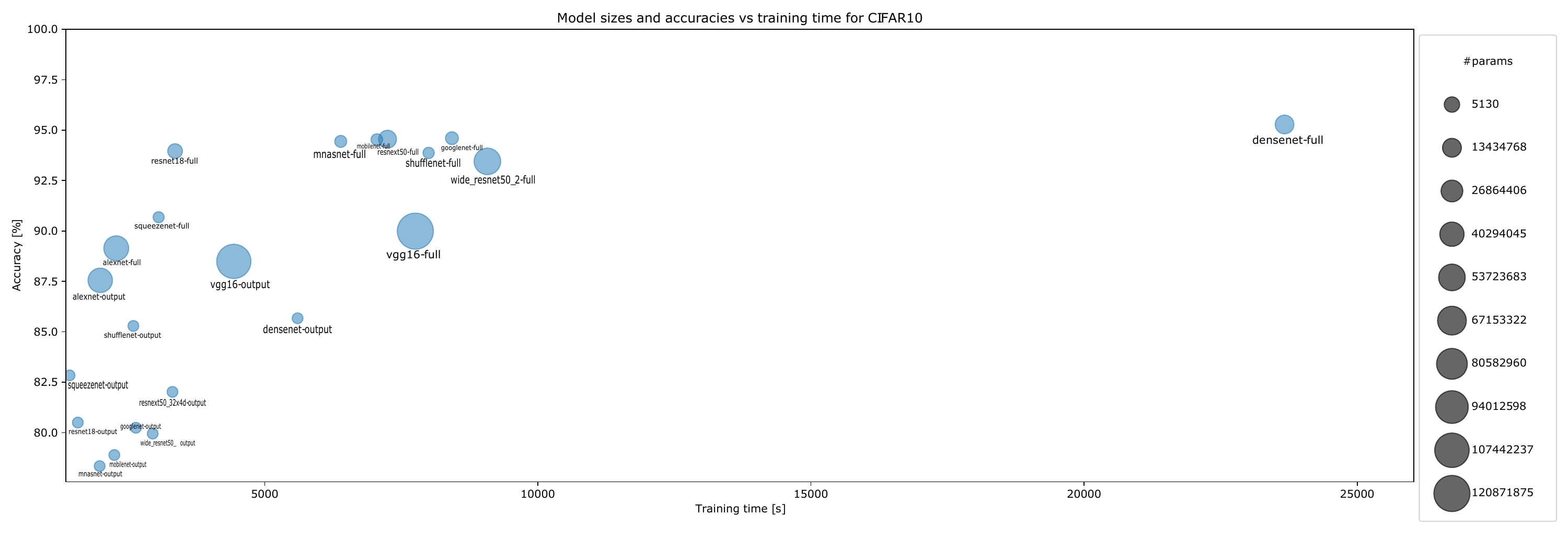}
\caption{\hl{Accuracy vs. training time} 
 and model size for CIFAR-10 for ten-episode~training.}
\label{10ep_scatter_cifar 10}
\end{figure}

\begin{figure}[H]
\includegraphics[width=\textwidth]{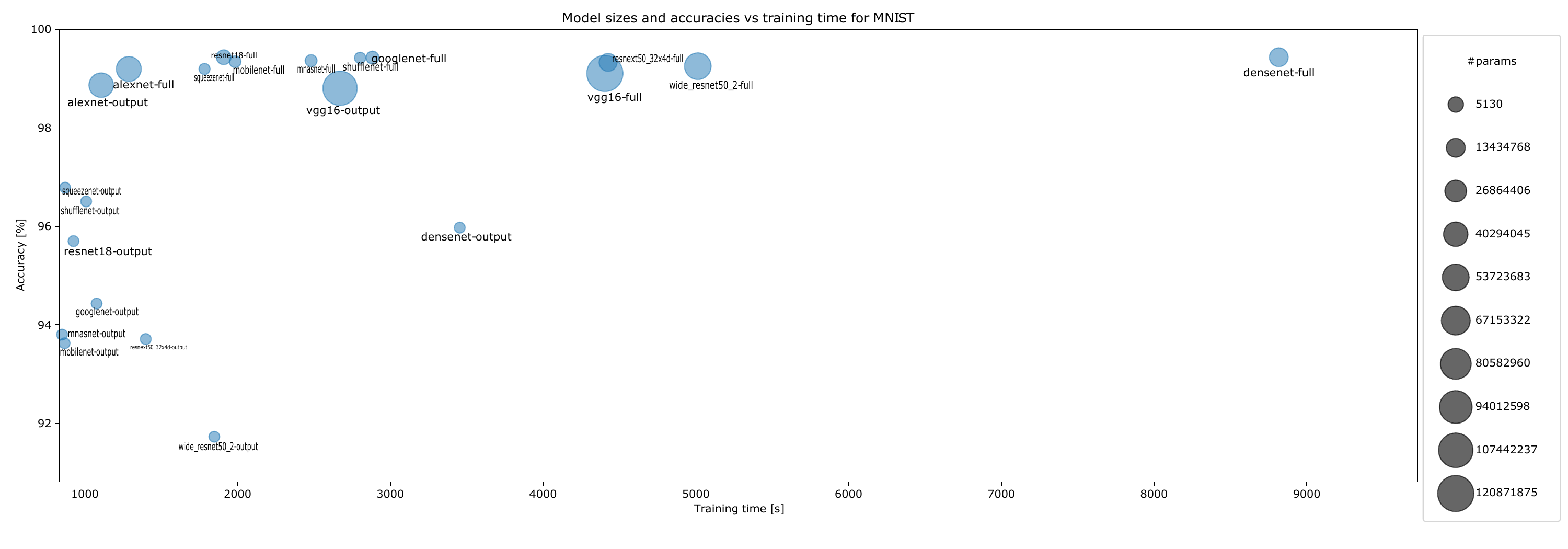}
\caption{Accuracy vs. training time and model size for MNIST for ten-episode~training.}
\label{10ep_scatter_mnist}
\end{figure}

\begin{figure}[H]
\includegraphics[width=\textwidth]{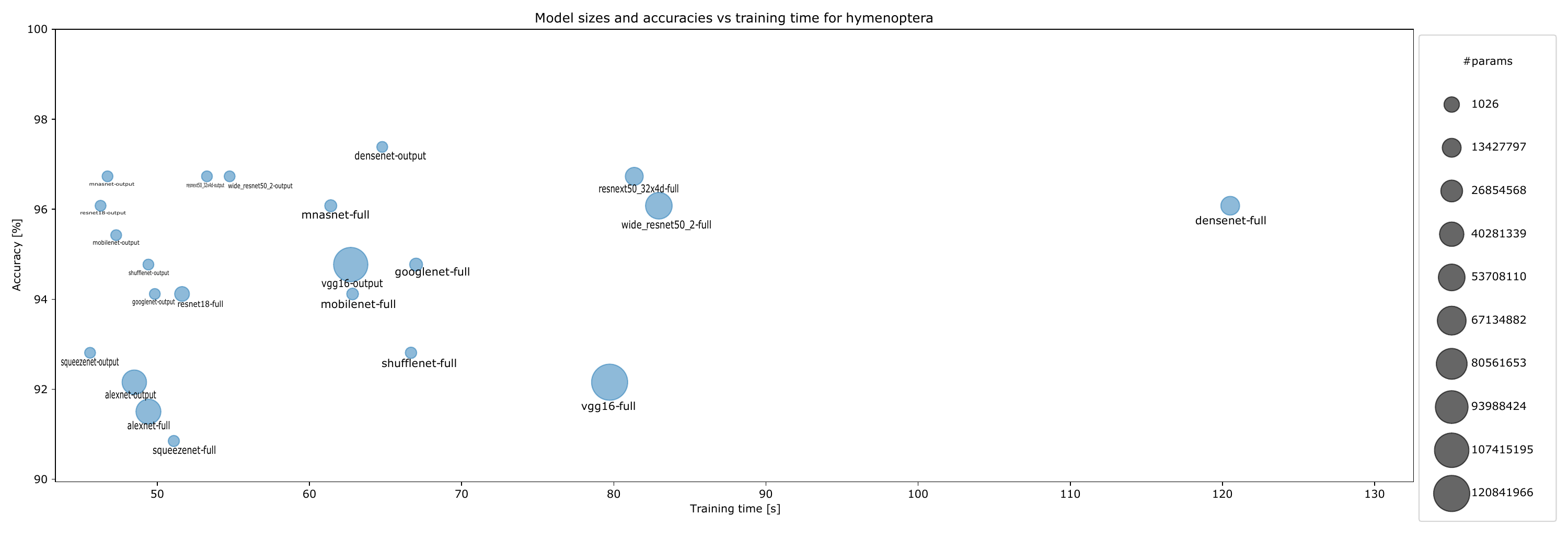}
\caption{Accuracy vs. training time and model size for Hymenoptera for ten-episode~training.}
\label{10ep_scatter_hymenoptera}
\end{figure}

\begin{figure}[H]
\includegraphics[width=\textwidth]{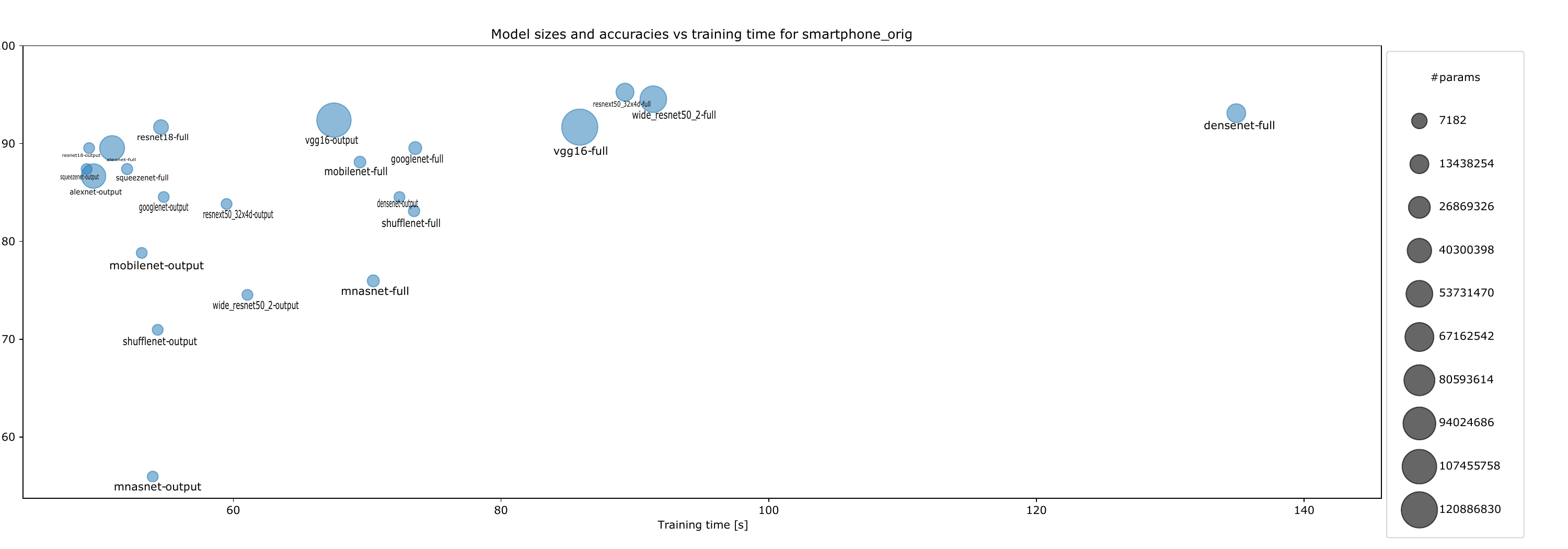}
\caption{Accuracy vs. training time and model size for original smartphone data for ten-episode~training.}
\label{10ep_scatter_smartphone_orig}
\end{figure}

\begin{figure}[H]
\includegraphics[width=\textwidth]{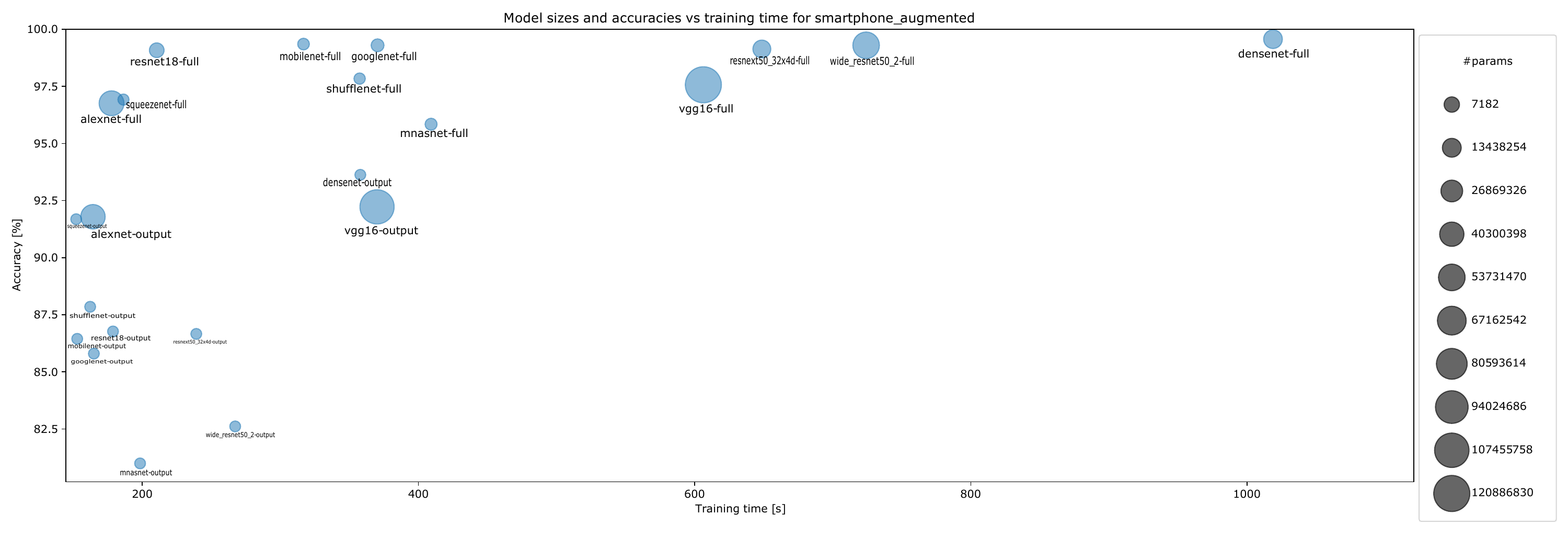}
\caption{Accuracy vs. training time and model size for augmented smartphone data for ten-episode~training.}
\label{10ep_scatter_smartphone_augmented}
\end{figure}

\begin{figure}[H]
\includegraphics[width=\textwidth, height = 0.32\textheight]{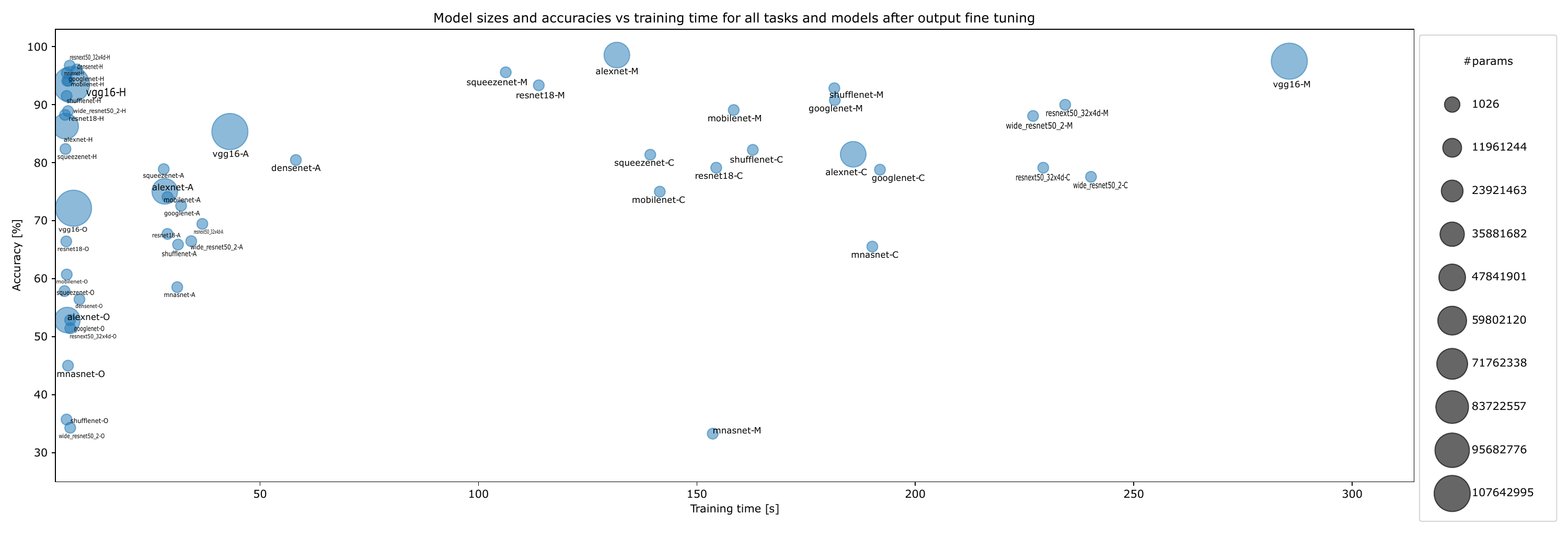}
\caption{Model sizes and accuracy vs. training time for all tasks and models after fine-tuning classifier layer only, where A refers to Augmented smartphones, C to CIFAR10, H to Hymenoptera, M to MNIST, and~O to the Original smartphone~dataset.\label{ASTOut}}
\end{figure}

\begin{figure}[H]
\includegraphics[width=\textwidth, height = 0.32\textheight]{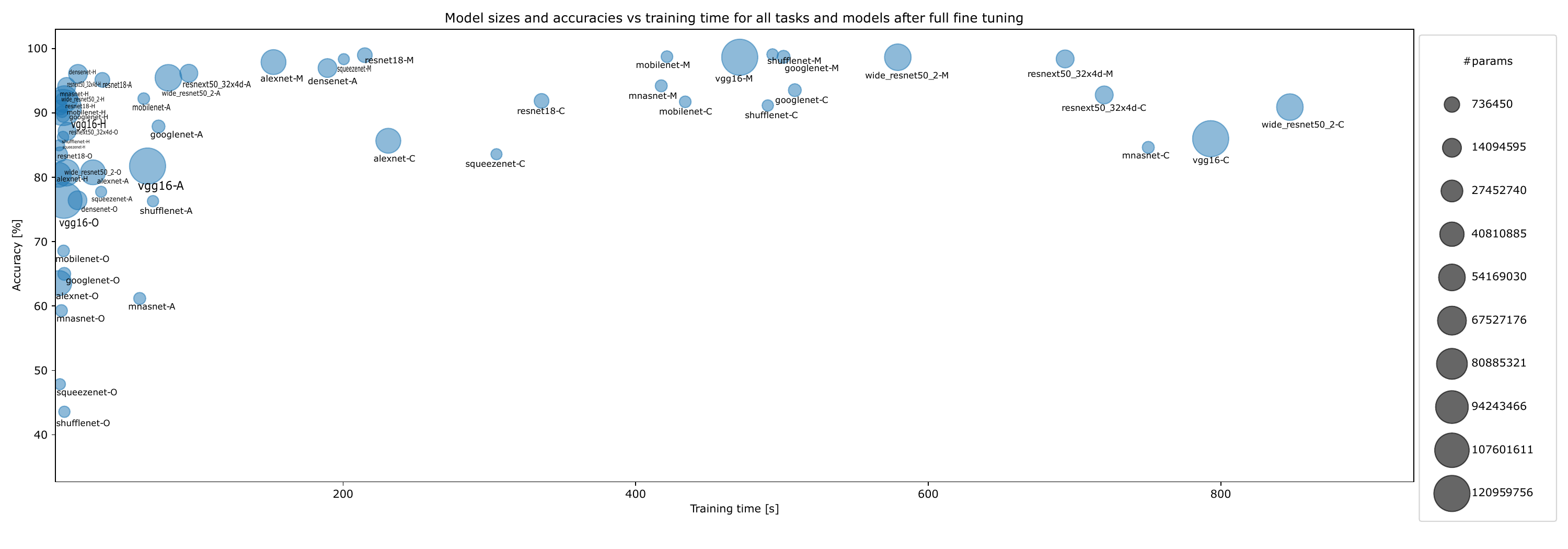}
\caption{Model sizes and accuracy vs. training time for all tasks and models after full fine-tuning, where A refers to Augmented smartphones, C to CIFAR10, H to Hymenoptera, M to MNIST, and~O to the Original smartphone~dataset.\label{ASTFull}}
\end{figure}

\begin{itemize}
\item Accuracy vs. training time and model size for CIFAR-10 for ten-episode training, Figure~\ref{10ep_scatter_cifar 10};
 \item \textls[-15]{Accuracy vs. training time and model size for MNIST for ten-episode training, \mbox{Figure~\ref{10ep_scatter_mnist}};}
\item Accuracy vs. training time and model size for Hymenoptera for ten-episode training, Figure \ref{10ep_scatter_hymenoptera};
\item Accuracy vs. training time and model size for original smartphone data for ten-episode training, Figure \ref{10ep_scatter_smartphone_orig};
\item Accuracy vs. training time and model size for augmented smartphone data for ten-episode training, Figure \ref{10ep_scatter_smartphone_augmented};
\item Model sizes and accuracy vs. training time for all tasks and models after fine-tuning the classifier layer only, where A refers to Augmented smartphones, C to CIFAR10, H to Hymenoptera, M to MNIST, and~O to the Original smartphone dataset, Figure~\ref{ASTOut};
\item Model sizes and accuracy vs. training time for all tasks and models after full fine-tuning, where A refers to Augmented smartphones, C to CIFAR10, H to Hymenoptera, M to MNIST, and~O to the Original smartphone dataset, Figure \ref{ASTFull}.
\end{itemize}

\unskip

\unskip

\unskip

\unskip

\unskip

\unskip

\section[\appendixname~\thesection]{Accuracy vs. Training Time and Model Size for Each Task with One-Episode~Learning}\label{AppAccModOne}
\begin{figure}[H]
\includegraphics[width=\textwidth]{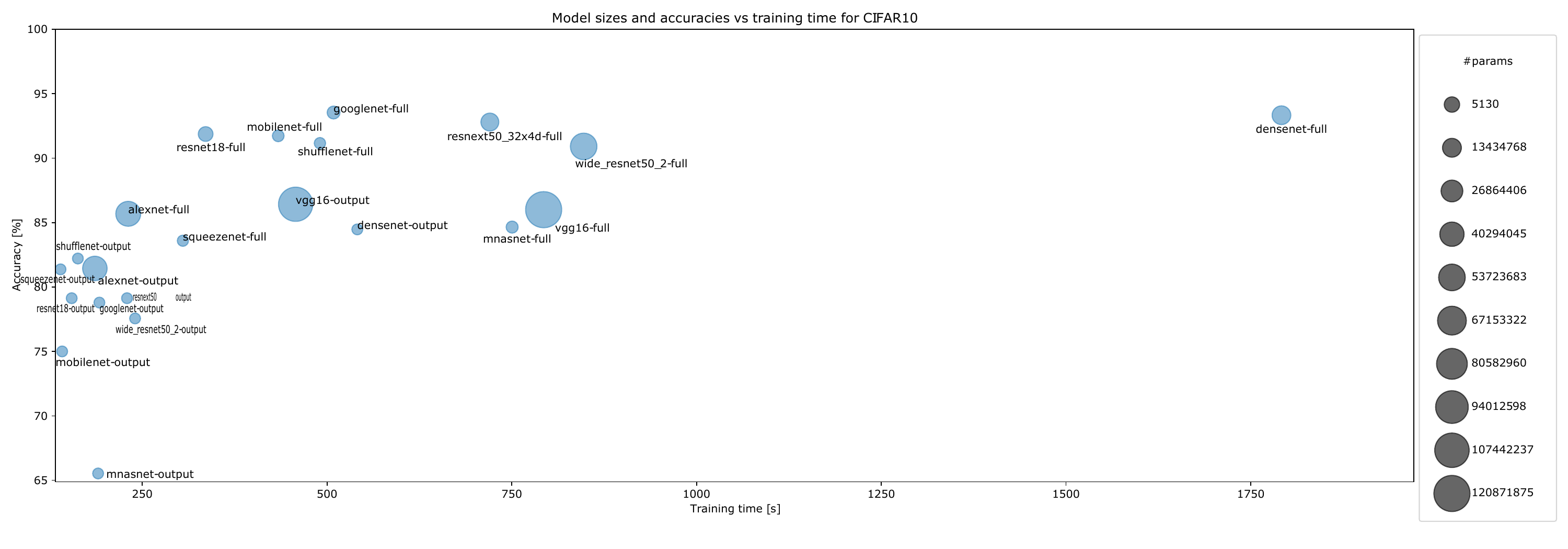}
\caption{Accuracy vs, training time and model size for CIFAR-10 for one-episode~training.}
\label{oneshot_scatter_cifar 10}
\end{figure}

\begin{figure}[H]
\includegraphics[width=\textwidth]{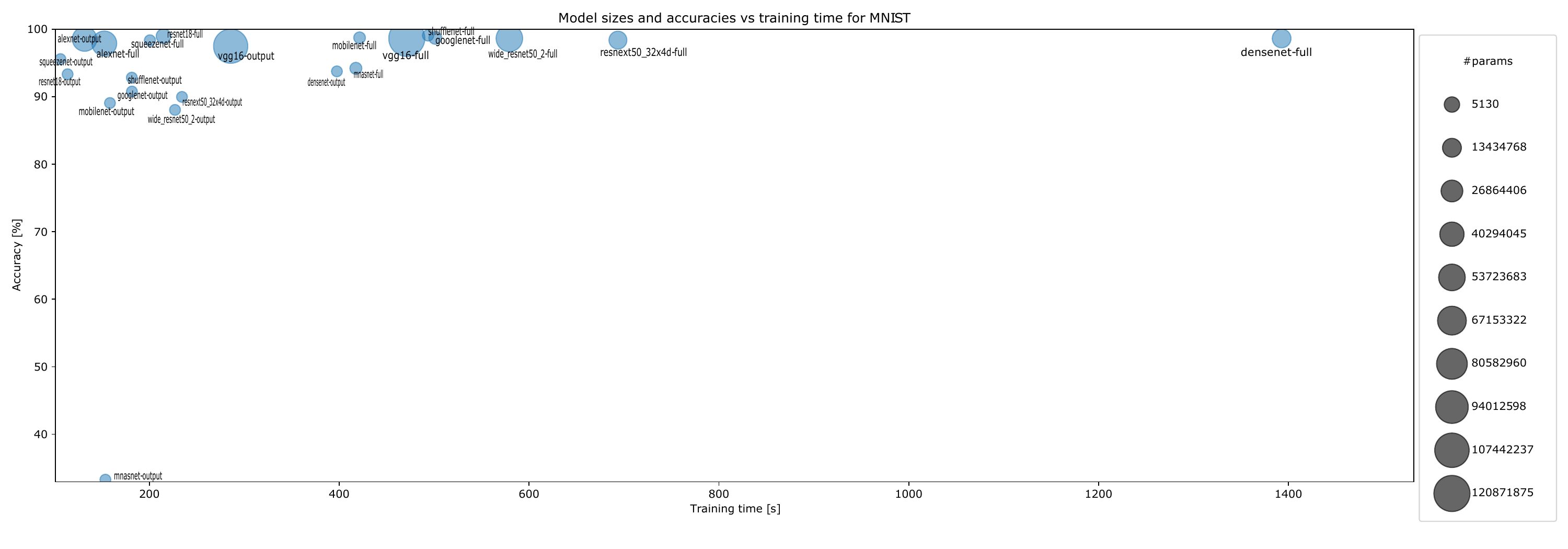}
\caption{Accuracy vs, training time and model size for MNIST for one-episode~training.}
\label{oneshot_scatter_mnist}
\end{figure}

\begin{figure}[H]
\includegraphics[width=\textwidth]{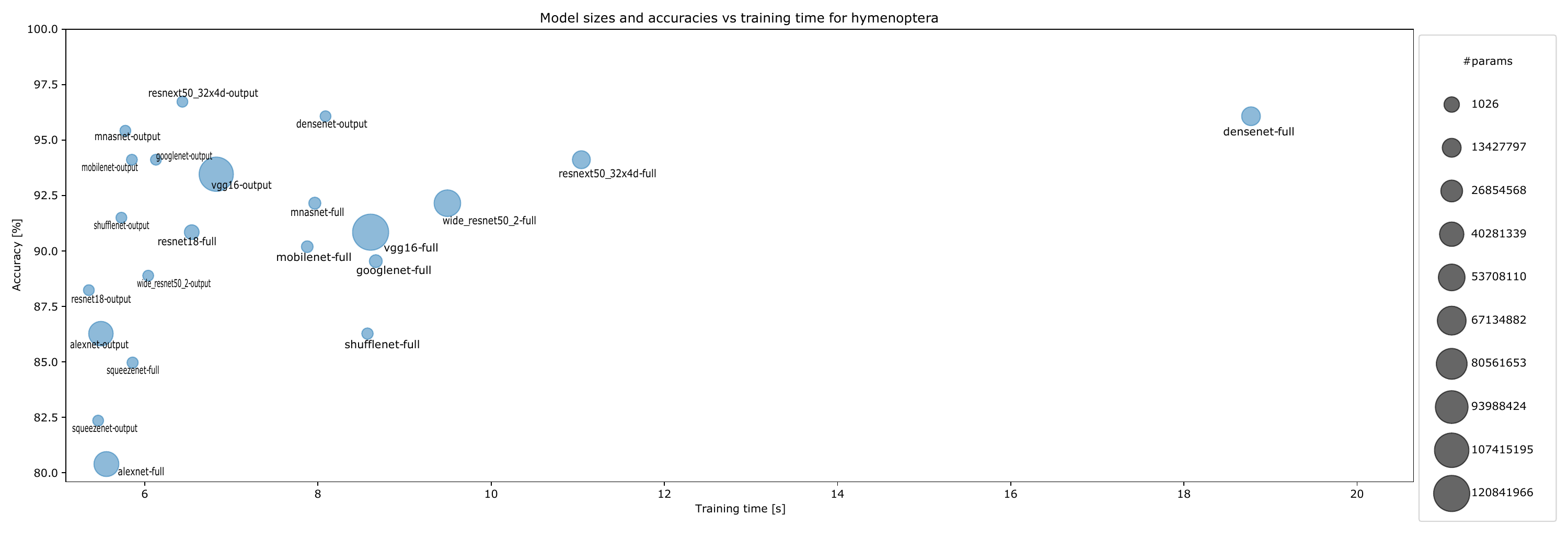}
\caption{Accuracy vs, training time and model size for Hymenoptera for one-episode~training.}
\label{oneshot_scatter_hymenoptera}
\end{figure}

\begin{figure}[H]
\includegraphics[width=\textwidth]{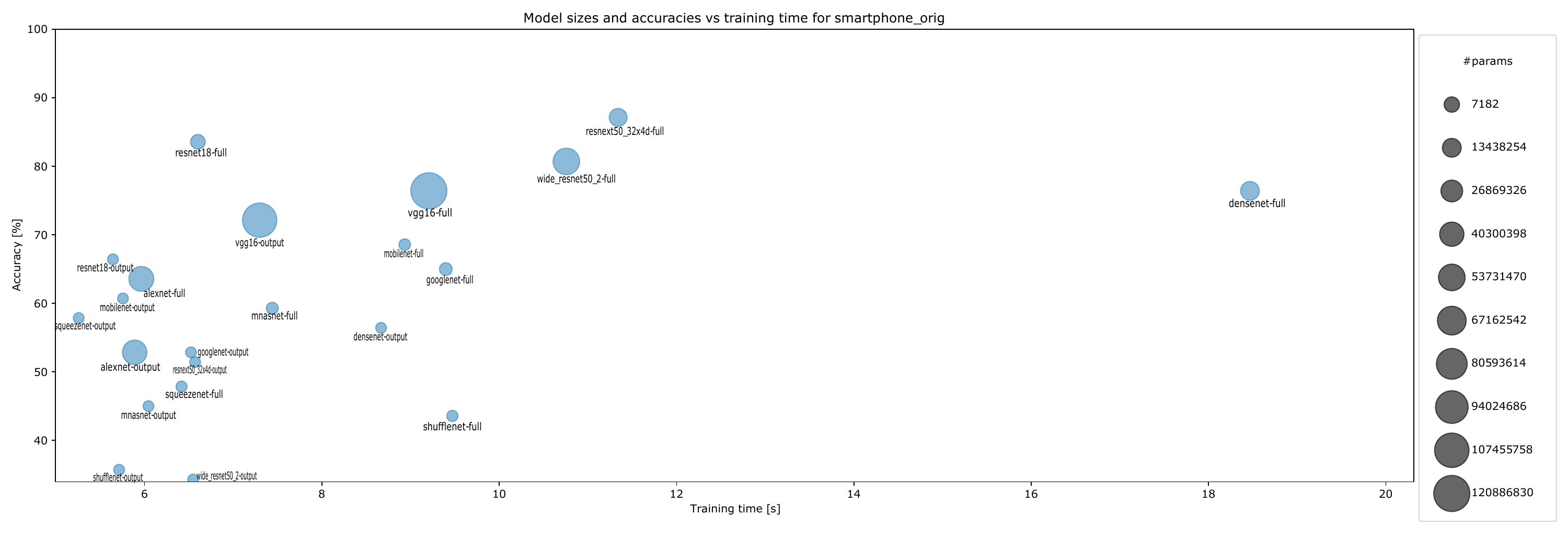}
\caption{Accuracy vs, training time and model size for original smartphone data for one-episode~training.}
\label{oneshot_scatter_smartphone_orig}
\end{figure}

\begin{figure}[H]
\includegraphics[width=\textwidth]{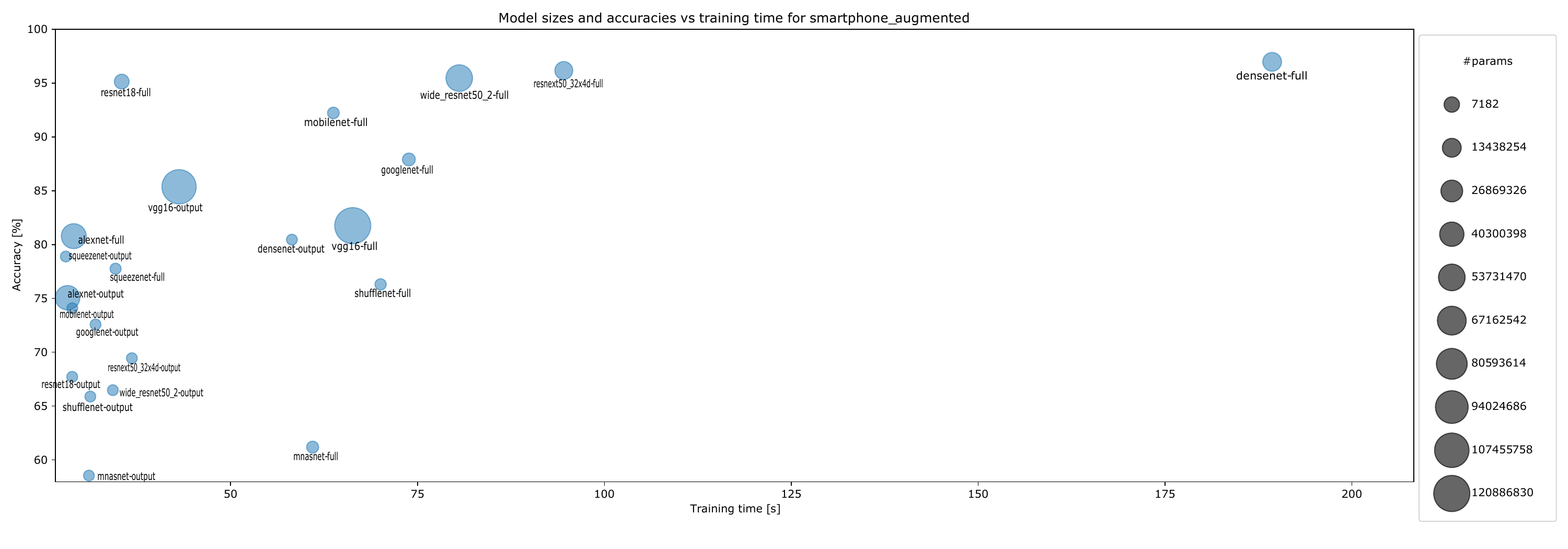}
\caption{Accuracy vs, training time and model size for augmented smartphone data for one-episode~training.}
\label{oneshot_scatter_smartphone_augmented}
\end{figure}

\begin{itemize}
\item Accuracy vs. training time and model size for CIFAR-10 for one-episode training, Figure~\ref{oneshot_scatter_cifar 10};
 \item \textls[-15]{ Accuracy vs. training time and model size for MNIST for one-episode training, \mbox{Figure~\ref{oneshot_scatter_mnist}};}
\item Accuracy vs. training time and model size for Hymenoptera for one-episode training, Figure \ref{oneshot_scatter_hymenoptera};
\item Accuracy vs. training time and model size for original smartphone data for one-episode training, Figure \ref{oneshot_scatter_smartphone_orig};
\item Accuracy vs. training time and model size for augmented smartphone data for one-episode training, Figure 
 \ref{oneshot_scatter_smartphone_augmented}.
\end{itemize}

\unskip

\unskip

\unskip

\unskip

\begin{adjustwidth}{-\extralength}{0cm}
\reftitle{References}

\end{adjustwidth}

%

\begin{thebibliography}{999}

\bibitem[Lundervold and Lundervold(2019)]{Lundervold.2019}
Lundervold, A.S.; Lundervold, A.
\newblock An overview of deep learning in medical imaging focusing on MRI.
\newblock {\em Z. Fur Med. Phys.} {\bf 2019}, {\em
 29},~102--127, doi:10.1016/j.zemedi.2018.11.002.

\bibitem[{Pires de Lima} and Marfurt(2020)]{PiresdeLima.2020}
{Pires de Lima}, R.; Marfurt, K.
\newblock Convolutional Neural Network for Remote-Sensing Scene Classification:
 Transfer Learning Analysis.
\newblock {\em Remote Sens.} {\bf 2020}, {\em 12},~86, doi:10.3390/rs12010086.

\bibitem[Zou and Zhong(2018)]{Zou.2018}
Zou, M.; Zhong, Y.
\newblock Transfer Learning for Classification of Optical Satellite Image.
\newblock {\em Sens. Imaging} {\bf 2018}, {\em 19}, doi:10.1007/s11220-018-0191-1.

\bibitem[Abou~Baker \em{et~al.}(2020)Abou~Baker, Szabo-M\"uller, and
 Handmann]{S-Cube-BakSzaHan2020}
Abou~Baker, N.; Szabo-M\"uller, P.; Handmann, U.
\newblock {Feature-fusion transfer learning method as a basis to support
 automated smartphone recycling in a circular smart city}.
\newblock In Proceedings of the EAI S-CUBE 2020---11th EAI International Conference on Sensor
 Systems and Software, Aalborg, Denmark, 10--11 December 2020.

\bibitem[Houlsby \em{et~al.}(2019)Houlsby, Giurgiu, Jastrzebski, Morrone,
 de~Laroussilhe, Gesmundo, Attariyan, and Gelly]{Houlsby.222019}
Houlsby, N.; Giurgiu, A.; Jastrzebski, S.; Morrone, B.; de~Laroussilhe, Q.;
 Gesmundo, A.; Attariyan, M.; Gelly, S.
\newblock Parameter-Efficient Transfer Learning for NLP. \emph{arXiv }\textbf{2019}, arXiv:1902.00751.

\bibitem[Choe \em{et~al.}(2020)Choe, Choi, and Kim]{Choe.2020}
Choe, D.; Choi, E.; Kim, D.K.
\newblock The Real-Time Mobile Application for Classifying of Endangered Parrot
 Species Using the CNN Models Based on Transfer Learning.
\newblock {\em Mob. Inf. Syst.} {\bf 2020}, {\em 2020},~1--13, doi:10.1155/2020/1475164.

\bibitem[Ismail~Fawaz \em{et~al.}(2018)Ismail~Fawaz, Forestier, Weber,
 Idoumghar, and Muller]{DBLP:journals/corr/abs-1811-01533}
Ismail~Fawaz, H.; Forestier, G.; Weber, J.; Idoumghar, L.; Muller, P.A.
\newblock Transfer learning for time series classification.
\newblock {In Proceedings of the 2018 IEEE International Conference on Big Data (Big Data)}, Seattle, WA, USA, 10--13 December {2018}; doi:10.1109/bigdata.2018.8621990. 

\bibitem[Canziani \em{et~al.}(2017)Canziani, Paszke, and
 Culurciello]{Canziani.2016}
Canziani, A.; Paszke, A.; Culurciello, E.
\newblock An Analysis of Deep Neural Network Models for Practical Applications. \emph{arXiv}
 \textbf{2017}, arXiv:1605.07678.

\bibitem[Bianco \em{et~al.}(2018)Bianco, Cadene, Celona, and
 Napoletano]{Bianco.2018}
Bianco, S.; Cadene, R.; Celona, L.; Napoletano, P.
\newblock Benchmark Analysis of Representative Deep Neural Network
 Architectures.
\newblock {\em IEEE Access} {\bf 2018}, {\em 6},~64270–64277, doi:10.1109/access.2018.2877890.

\bibitem[Socher \em{et~al.}(2013)Socher, Ganjoo, Sridhar, Bastani, Manning, and
 Ng]{socher2013zeroshot}
Socher, R.; Ganjoo, M.; Sridhar, H.; Bastani, O.; Manning, C.D.; Ng, A.Y.
\newblock Zero-Shot Learning Through Cross-Modal Transfer. \emph{arXiv} \textbf{2013}, arXiv:1301.3666.

\bibitem[Xian \em{et~al.}(2020)Xian, Schiele, and Akata]{xian2020zeroshot}
Xian, Y.; Schiele, B.; Akata, Z.
\newblock Zero-Shot Learning---The Good, the Bad and the Ugly. \emph{arXiv} \textbf{2020}, arXiv:1703.04394.

\bibitem[Lampert \em{et~al.}(2014)Lampert, Nickisch, and Harmeling]{6571196}
Lampert, C.H.; Nickisch, H.; Harmeling, S.
\newblock Attribute-Based Classification for Zero-Shot Visual Object
 Categorization.
\newblock {\em IEEE Trans. Pattern Anal. Mach. Intell.}
 {\bf 2014}, {\em 36},~453--465, doi:10.1109/TPAMI.2013.140.

\bibitem[Zhang and Saligrama(2015)]{zhang2015zeroshot}
Zhang, Z.; Saligrama, V.
\newblock Zero-Shot Learning via Semantic Similarity Embedding. \emph{arXiv} \textbf{2015}, arXiv:1509.04767.

\bibitem[Akata \em{et~al.}(2016)Akata, Perronnin, Harchaoui, and Schmid]{2016}
Akata, Z.; Perronnin, F.; Harchaoui, Z.; Schmid, C.
\newblock Label-Embedding for Image Classification.
\newblock {\em IEEE Trans. Pattern Anal. Mach. Intell.}
 {\bf 2016}, {\em 38},~1425–1438, doi:10.1109/tpami.2015.2487986.

\bibitem[Bart and Ullman(2005)]{1467333}
Bart, E.; Ullman, S.
\newblock Cross-generalization: Learning novel classes from a single example by
 feature replacement.
\newblock In Proceedings of the 2005 IEEE Computer Society Conference on Computer Vision and Pattern
 Recognition (CVPR'05), San Diego, CA, USA, 20--26 June 2005; Volume~1, pp. 672--679, doi:10.1109/CVPR.2005.117.

\bibitem[Fink(2005)]{NIPS2004_ef1e491a}
Fink, M.
\newblock Object Classification from a Single Example Utilizing Class Relevance
 Metrics.
\newblock In \emph{Advances in Neural Information Processing Systems}; Saul, L., Weiss,
 Y., Bottou, L., Eds.; MIT Press: Cambridge, MA, USA, 2005; Volume~17.

\bibitem[Tommasi and Caputo(2009)]{Tommasi2009TheMY}
Tommasi, T.; Caputo, B.
\newblock The More You Know, the Less You Learn: From Knowledge Transfer to
 One-shot Learning of Object Categories.
\newblock In Proceedings of the BMVC, 2009. Available online: \url{http://www.bmva.org/bmvc/2009/Papers/Paper353/Paper353.html} ({accessed on 30 November 2021}
).

\bibitem[Wang \em{et~al.}(2020)Wang, Yao, Kwok, and Ni]{10.1145/3386252}
Wang, Y.; Yao, Q.; Kwok, J.T.; Ni, L.M.
\newblock Generalizing from a Few Examples: A Survey on Few-Shot Learning.
\newblock {\em ACM Comput. Surv.} {\bf 2020}, {\em 53}, doi:10.1145/3386252.

\bibitem[Azadi \em{et~al.}(2017)Azadi, Fisher, Kim, Wang, Shechtman, and
 Darrell]{azadi2017multicontent}
Azadi, S.; Fisher, M.; Kim, V.; Wang, Z.; Shechtman, E.; Darrell, T.
\newblock Multi-Content GAN for Few-Shot Font Style Transfer. \emph{arXiv} \textbf{2017}, arXiv:1712.00516.

\bibitem[Liu \em{et~al.}(2019)Liu, Wang, Dixit, Kwitt, and
 Vasconcelos]{liu2019feature}
Liu, B.; Wang, X.; Dixit, M.; Kwitt, R.; Vasconcelos, N.
\newblock Feature Space Transfer for Data Augmentation. \emph{arXiv} \textbf{2019}, arXiv:1801.04356.

\bibitem[Luo \em{et~al.}(2017)Luo, Zou, Hoffman, and Fei-Fei]{luo2017label}
Luo, Z.; Zou, Y.; Hoffman, J.; Fei-Fei, L.
\newblock Label Efficient Learning of Transferable Representations across
 Domains and Tasks. \emph{arXiv} \textbf{2017}, arXiv:1712.00123.

\bibitem[Tan \em{et~al.}(2018)Tan, Chen, Pantazis, and Pan]{8560550}
Tan, W.C.; Chen, I.M.; Pantazis, D.; Pan, S.J.
\newblock Transfer Learning with PipNet: For Automated Visual Analysis of
 Piping Design.
\newblock In Proceedings of the 2018 IEEE 14th International Conference on Automation Science and
 Engineering (CASE), Munich, Germany, 20--24 August 2018; pp. 1296--1301, doi:10.1109/COASE.2018.8560550.

\bibitem[Montúfar \em{et~al.}(2014)Montúfar, Pascanu, Cho, and
 Bengio]{Montufar.282014}
Montúfar, G.; Pascanu, R.; Cho, K.; Bengio, Y.
\newblock On the Number of Linear Regions of Deep Neural Networks. \emph{arXiv} \textbf{2014}, arXiv:1402.1869.

\bibitem[Kawaguchi \em{et~al.}(2019)Kawaguchi, Huang, and
 Kaelbling]{Kawaguchi.2019}
Kawaguchi, K.; Huang, J.; Kaelbling, L.P.
\newblock Effect of Depth and Width on Local Minima in Deep Learning.
\newblock {\em Neural Comput.} {\bf 2019}, {\em 31},~1462–1498, doi:10.1162/neco\_a\_01195.

\bibitem[Khan \em{et~al.}(2020)Khan, Sohail, Zahoora, and Qureshi]{Khan.2020}
Khan, A.; Sohail, A.; Zahoora, U.; Qureshi, A.S.
\newblock A survey of the recent architectures of deep convolutional neural
 networks.
\newblock {\em Artif. Intell. Rev.} {\bf 2020}, {\em
 53},~5455--5516, doi:10.1007/s10462-020-09825-6.

\bibitem[Hochreiter(1998)]{Hochreiter.1998}
Hochreiter, S.
\newblock The Vanishing Gradient Problem during Learning Recurrent Neural Nets
 and Problem Solutions.
\newblock {\em Int. J. Uncertain. Fuzziness Knowl.-Based Syst.} {\bf 1998},
 {\em 6},~107–116, doi:10.1142/S0218488598000094.

\bibitem[Srivastava \em{et~al.}(2015)Srivastava, Greff, and
 Schmidhuber]{Srivastava.532015}
Srivastava, R.K.; Greff, K.; Schmidhuber, J.
\newblock Highway Networks. \emph{arXiv} \textbf{2015}, arXiv:1505.00387.

\bibitem[Hu \em{et~al.}(2018)Hu, Shen, and Sun]{Hu.2018}
Hu, J.; Shen, L.; Sun, G.
\newblock Squeeze-and-Excitation Networks.
\newblock In Proceedings of the 2018 IEEE/CVF Conference on Computer Vision and Pattern Recognition, Salt Lake City, UT, USA, 18--22 June 2018; pp. 7132--7141, doi:10.1109/CVPR.2018.00745.

\bibitem[{Alex Krizhevsky} \em{et~al.}(2012){Alex Krizhevsky}, {Ilya
 Sutskever}, and {Geoffrey E. Hinton}]{AlexKrizhevsky.2012}
{Krizhevsky}, A.; { Sutskever}, I.; {Hinton}, G.E.
\newblock ImageNet Classification with Deep Convolutional Neural  Networks.
\newblock {\em Adv. Neural Inf. Process. Syst.} {\bf 2012},{ \em 25},~1097–1105. 

\bibitem[Simonyan and Zisserman()]{Simonyan.942014}
Simonyan, K.; Zisserman, A.
\newblock Very Deep Convolutional Networks for Large-Scale Image Recognition. \emph{arXiv} \textbf{{2014}}, {arXiv:1409.1556}. 

\bibitem[Szegedy \em{et~al.}(6/7/2015 - 6/12/2015)Szegedy, Liu, Jia, Sermanet,
 Reed, Anguelov, Erhan, Vanhoucke, and Rabinovich]{Szegedy.6720156122015}
Szegedy, C.; Liu, W.; Jia, Y.; Sermanet, P.; Reed, S.; Anguelov, D.; Erhan, D.;
 Vanhoucke, V.; Rabinovich, A.
\newblock Going deeper with convolutions.
\newblock In Proceedings of the 2015 IEEE Conference on Computer Vision and Pattern Recognition
 (CVPR), Boston, MA, USA, 7--12 June 2015; pp. 1--9, doi:10.1109/CVPR.2015.7298594.

\bibitem[He \em{et~al.}(2015)He, Zhang, Ren, and Sun]{He.12102015}
He, K.; Zhang, X.; Ren, S.; Sun, J.
\newblock Deep Residual Learning for Image Recognition. \emph{arXiv} \textbf{2015}, arXiv:1512.03385.

\bibitem[Iandola \em{et~al.}(2016)Iandola, Han, Moskewicz, Ashraf, Dally, and
 Keutzer]{Iandola.2242016}
Iandola, F.N.; Han, S.; Moskewicz, M.W.; Ashraf, K.; Dally, W.J.; Keutzer, K.
\newblock SqueezeNet: AlexNet-level accuracy with 50x fewer parameters and
 <0.5MB model size. \emph{arXiv} \textbf{2016}, arXiv:1602.07360.

\bibitem[Xie \em{et~al.}(7/21/2017 - 7/26/2017)Xie, Girshick, Dollar, Tu, and
 He]{Xie.72120177262017}
Xie, S.; Girshick, R.; Dollar, P.; Tu, Z.; He, K.
\newblock Aggregated Residual Transformations for Deep Neural Networks.
\newblock In Proceedings of the 2017 IEEE Conference on Computer Vision and Pattern Recognition
 (CVPR), Honolulu, HI, USA, 21--26 July 2017; pp.~5987--5995, doi:10.1109/CVPR.2017.634.

\bibitem[Huang \em{et~al.}(7/21/2017 - 7/26/2017)Huang, Liu, {van der Maaten},
 and Weinberger]{Huang.72120177262017}
Huang, G.; Liu, Z.; {van der Maaten}, L.; Weinberger, K.Q.
\newblock Densely Connected Convolutional Networks.
\newblock In Proceedings of the 2017 IEEE Conference on Computer Vision and Pattern Recognition
 (CVPR), Honolulu, HI, USA, 21--26 July 2017; pp. 2261--2269, doi:10.1109/CVPR.2017.243.

\bibitem[Howard \em{et~al.}(2017)Howard, Zhu, Chen, Kalenichenko, Wang, Weyand,
 Andreetto, and Adam]{Howard.4172017}
Howard, A.G.; Zhu, M.; Chen, B.; Kalenichenko, D.; Wang, W.; Weyand, T.;
 Andreetto, M.; Adam, H.
\newblock MobileNets: Efficient Convolutional Neural Networks for Mobile Vision
 Applications. \emph{arXiv} \textbf{2017}, arXiv:1704.04861.

\bibitem[Zagoruyko and Komodakis(2017)]{Zagoruyko.5232016}
Zagoruyko, S.; Komodakis, N.
\newblock Wide Residual Networks. \emph{arXiv} \textbf{2017}, arXiv:1605.07146.

\bibitem[Zhang \em{et~al.}(2017)Zhang, Zhou, Lin, and Sun]{Zhang.742017}
Zhang, X.; Zhou, X.; Lin, M.; Sun, J.
\newblock ShuffleNet: An Extremely Efficient Convolutional Neural Network for
 Mobile Devices. \emph{arXiv} \textbf{2017}, arXiv:1707.01083.

\bibitem[Tan \em{et~al.}(2019)Tan, Chen, Pang, Vasudevan, Sandler, Howard, and
 Le]{Tan}
Tan, M.; Chen, B.; Pang, R.; Vasudevan, V.; Sandler, M.; Howard, A.; Le, Q.V.
\newblock MnasNet: Platform-Aware Neural Architecture Search for Mobile. \emph{arXiv} \textbf{2019}, arXiv:1807.11626

\bibitem[Zaheer and Shaziya(2019)]{Zaheer}
Zaheer, R.; Shaziya, H.
\newblock A Study of the Optimization Algorithms in Deep Learning.
\newblock In Proceedings of the 2019 Third International Conference on Inventive Systems and Control
 (ICISC), Coimbatore, India, 10--11 January 2019; pp. 536--539, doi:10.1109/ICISC44355.2019.9036442.

\bibitem[Kaziha and Bonny(11/19/2019 - 11/21/2019)]{Kaziha.1119201911212019}
Kaziha, O.; Bonny, T.
\newblock A Comparison of Quantized Convolutional and LSTM Recurrent Neural
 Network Models Using MNIST.
\newblock In Proceedings of the 2019 International Conference on Electrical and Computing
 Technologies and Applications (ICECTA), Ras Al Khaimah, United Arab Emirates, 19--21 November 2019; pp.
 1--5, doi:10.1109/ICECTA48151.2019.8959793.

\bibitem[Paszke \em{et~al.}(2019)Paszke, Gross, Massa, Lerer, Bradbury, Chanan,
 Killeen, Lin, Gimelshein, Antiga, Desmaison, Kopf, Yang, DeVito, Raison,
 Tejani, Chilamkurthy, Steiner, Fang, Bai, and Chintala]{NEURIPS2019_9015}
Paszke, A.; Gross, S.; Massa, F.; Lerer, A.; Bradbury, J.; Chanan, G.; Killeen,
 T.; Lin, Z.; Gimelshein, N.; Antiga, L.; et al.
\newblock PyTorch: An Imperative Style, High-Performance Deep Learning Library.
\newblock {\em Adv. Neural Inf. Process. Syst.} {\bf 2019}, {\em 32},~8024–8035.

\bibitem[Baker \em{et~al.}(2021)Baker, Szabo-M\'{y}ller, and
 Handmann]{smartphones.EAI}
Baker, N.A.; Szabo-M\'{y}ller, P.; Handmann, U.
\newblock Transfer learning-based method for automated e-waste recycling in
 smart cities.
\newblock {\em EAI Endorsed Trans. Smart Cities} {\bf 2021}, {\em 5}, doi:10.4108/eai.16-4-2021.169337.

\bibitem[Chen \em{et~al.}(2021)Chen, Li, Bai, Yang, Jiang, and
 Miao]{rs13224712}
Chen, L.; Li, S.; Bai, Q.; Yang, J.; Jiang, S.; Miao, Y.
\newblock Review of Image Classification Algorithms Based on Convolutional
 Neural Networks.
\newblock {\em Remote Sens.} {\bf 2021}, {\em 13}, 4712, doi:10.3390/rs13224712.

\end{thebibliography}



\end{document}